\documentclass{article} 
\usepackage{iclr2024_conference,times}

\usepackage{amsmath,amsfonts,bm}

\def\eqref#1{equation~\ref{#1}}

\def\1{\bm{1}}

\DeclareMathAlphabet{\mathsfit}{\encodingdefault}{\sfdefault}{m}{sl}
\SetMathAlphabet{\mathsfit}{bold}{\encodingdefault}{\sfdefault}{bx}{n}

\DeclareMathOperator*{\argmax}{arg\,max}

\usepackage{url}
\usepackage{xspace}
\usepackage{caption}
\usepackage{wrapfig}
\usepackage{algorithm}
\usepackage{algpseudocode}
\usepackage{amsmath,amssymb}
\usepackage{float}
\usepackage{graphicx}
\usepackage{csquotes}
\usepackage{xcolor}
\usepackage{multirow}
\usepackage{multicol}
\usepackage{wrapfig,lipsum,booktabs}
\usepackage{makecell}
\usepackage{wrapfig}

\usepackage{array}
\usepackage{color, colortbl}
\usepackage{pifont}
\usepackage{enumitem}
\usepackage{inconsolata}
\usepackage{subcaption}
\usepackage[hidelinks,colorlinks=true,linkcolor=blue,citecolor=blue]{hyperref}
\usepackage{xcolor}
\usepackage[normalem]{ulem} 
\usepackage{soul}

\definecolor{dkcolor}{RGB}{255,50,50}
\setulcolor{dkcolor}
\setul{0.4ex}{1.2pt}

\def\ours{\text{PromptAgent}\xspace}
\newcommand{\model}[1]{\texttt{#1}}

\usepackage{soul}
\newcommand{\ctext}[3][RGB]{%
  \begingroup
  \definecolor{hlcolor}{#1}{#2}\sethlcolor{hlcolor}%
  \hl{#3}%
  \endgroup
}

\title{PromptAgent: Strategic Planning with \\Language Models Enables Expert-level \\Prompt Optimization}

\author{Xinyuan Wang\textsuperscript{$1$}\thanks{Equal contribution} \ 
Chenxi Li\textsuperscript{$1*$} \ 
Zhen Wang\textsuperscript{$12*$}\thanks{Corresponding author} \ \
Fan Bai\textsuperscript{$5$} \  
Haotian Luo\textsuperscript{$2$}\\  
\textbf{Jiayou Zhang\textsuperscript{$2$} \  
Nebojsa Jojic\textsuperscript{$3$} \
Eric Xing\textsuperscript{$24$} \ 
Zhiting Hu\textsuperscript{$1$} } \\
\textsuperscript{$1$}UC San Diego 
\textsuperscript{$4$}Carnegie Mellon University \\
\textsuperscript{$3$}Microsoft Research 
\textsuperscript{$5$}Georgia Institute of Technology
\\
\textsuperscript{$2$}Mohamed bin Zayed University of Artificial Intelligence \\
\texttt{\{xiw136, chl078, zhw085, zhh019\}@ucsd.edu} 
}

\iclrfinalcopy 
\begin{document}

\maketitle

\begin{abstract}

Highly effective, task-specific prompts are often heavily engineered by experts to integrate detailed instructions and domain insights based on a deep understanding of both instincts of large language models (LLMs) and the intricacies of the target task. However, automating the generation of such expert-level prompts remains elusive. Existing prompt optimization methods tend to overlook the depth of domain knowledge and struggle to efficiently explore the vast space of expert-level prompts. Addressing this, we present {PromptAgent}, an optimization method that autonomously crafts prompts equivalent in quality to those handcrafted by experts. At its core, PromptAgent views prompt optimization as a strategic planning problem and employs a principled planning algorithm, rooted in Monte Carlo tree search, to strategically navigate the expert-level prompt space. Inspired by human-like trial-and-error exploration, PromptAgent induces precise expert-level insights and in-depth instructions by reflecting on model errors and generating constructive error feedback. Such a novel framework allows the agent to iteratively examine intermediate prompts (states), refine them based on error feedbacks (actions), simulate future rewards, and search for high-reward paths leading to expert prompts. We apply PromptAgent to 12 tasks spanning three practical domains: BIG-Bench Hard (BBH), as well as domain-specific and general NLP tasks, showing it significantly outperforms strong Chain-of-Thought and recent prompt optimization baselines. Extensive analyses emphasize its capability to craft expert-level, detailed, and domain-insightful prompts with great efficiency and generalizability\footnote{Code and demo are available at: \url{https://github.com/XinyuanWangCS/PromptAgent}}.

\end{abstract}

\vspace{-10pt}
\section{Introduction}
\vspace{-5pt}

\begin{figure}
    \centering
     \includegraphics[width=0.85\linewidth]{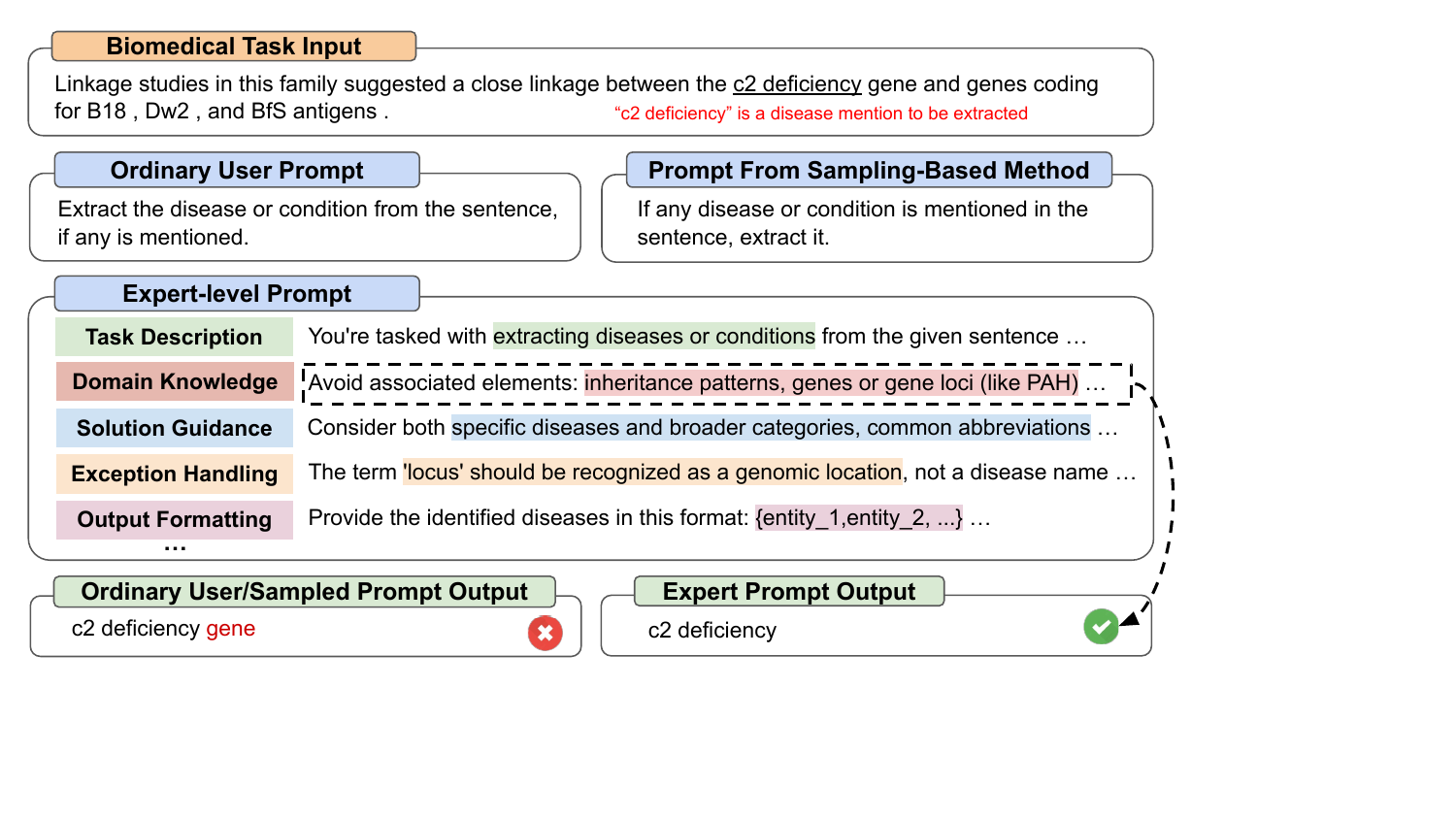}
    \vspace{-5pt}
    \caption{{Expert-level prompt} vs.\ ordinary human-written prompt and prompt from sampling-based methods (i.e., Automatic Prompt Engineer, \cite{zhou2022large}). The task is in the biomedical domain for extracting disease entities (NCBI, \citet{dougan2014ncbi}). The expert prompt provides much richer domain-specific details and structured guidance than the other two, leading to the correct prediction.}
    \vspace{-22pt}
    \label{fig:expert_prompt}
\end{figure}

Prompt engineering aims to craft effective prompts for harnessing the full potential of large language models (LLMs). Recent automatic prompt engineering, i.e., prompt optimization, has successfully studied training soft prompts~\citep{lester2021power, hu2021lora, wang2022multitask}, or searching for optimal combinations of discrete tokens~\citep{shin2020autoprompt, deng2022rlprompt, zhang2022tempera}, by utilizing internal states or gradients of LLMs. For cutting-edge, proprietary API-based LLMs like GPT-4~\citep{OpenAI2023GPT4TR}, prompt engineering largely relies on somewhat ad-hoc human-machine interactions. Human prompting experts thus need a unique blend of domain knowledge and intuition for LLMs to design the most effective prompts. For instance, an ideal prompt from human experts, shown in Figure~\ref{fig:expert_prompt}, might integrate nuanced elements like task descriptions, domain knowledge, solution guidance, etc., all of which substantially boost prompt quality and performance.

\begin{wrapfigure}{R}{0.3\textwidth}
\vspace{-15pt}
    \centering
    \includegraphics[width=0.28\textwidth]{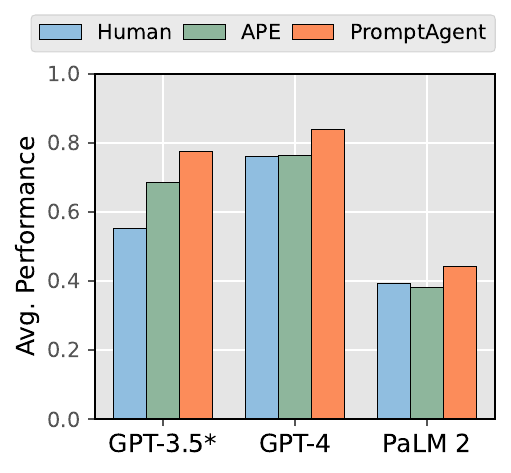}
    \vspace{-5pt}
    \caption{Prompt comparison across different base models. 
    }
    \vspace{-10pt}
    \label{fig:prompt_transfer_bar}
\end{wrapfigure}

Automating expert-level prompting engineering on API-based LLMs presents significant challenges, largely due to the intricate nature of expert-level prompts, as illustrated in Figure~\ref{fig:expert_prompt}. Although recent prompt optimization approaches have begun to utilize techniques like iterative sampling or evolutionary algorithms, such as Monte Carlo search~\citep{zhou2022large} or Gibbs sampling~\citep{xu2023reprompting}, they mostly employ heuristic methods like text edits or paraphrasing for generating candidate prompts~\citep{zhou2022large, prasad2023grips}. These approaches also often rely on straightforward iteration algorithms and lack a principled strategy to guide the exploration. Consequently, they tend to settle on local variants of prompts from ordinary users and rarely ascend to the excellence and nuances of expert-level prompts. Critically, many of these methods overlook that prompting engineering is essentially a human-in-the-loop application. In this process, humans refine prompts by fixing intermediate errors and integrating necessary domain knowledge through iterative interactions. This iterative refinement process characterizes the merits of how human experts craft superior prompts. Yet, the challenge remains that human exploration, while effective, can be expensive and less efficient at handling multiple errors simultaneously to explore the prompt space, thereby impeding the scalability of expert-level prompting.

To address the above challenges and combine human-like exploration with machine efficiency, we introduce \ours in this paper. Drawing inspiration from human trial-and-error processes, \ours seamlessly incorporates the principled planning approach, specifically Monte Carlo Tree Search (MCTS), to strategically optimize the prompting process. Notably, \ours reformulates prompt optimization as a strategic planning problem to address the complexity of expert-level prompt space. Under this planning framework, it plays trial-and-error iteration to retrieve model errors and leverages the self-reflection ability of LLMs~\citep{jang2023reflection, shinn2023reflexion, pan2023automatically} to generate insightful \textit{error feedback}. This feedback, in turn, plays a critical role in effectively inducing domain knowledge and guiding towards in-depth prompts. Through strategic planning, \ours iteratively leverages insightful error feedback (action) to refine each version of prompts (state). Starting from an initial prompt (state), \ours systematically grows the prompt space in a tree structure and prioritizes high-reward traces to navigate the vast space of expert-level prompts. The principled MCTS planning allows \ours to look ahead and simulate future rewards, which are then backpropagated to update the beliefs about the current prompt so that \ours can explore more promising alternatives later.

We demonstrate that \ours can discover productive expert-level prompts by applying it to 12 tasks spanning three practical and distinct domains: BIG-Bench Hard (BBH)~\citep{suzgun2022challenging}, as well as domain-specific and general NLP tasks. Starting with an initial human-written prompt and a small set of training samples, \ours not only enhances the performance of the initial human prompt greatly but also significantly surpasses strong Chain-of-Thought (CoT) and recent prompt optimization baselines. For instance, Figure~\ref{fig:prompt_transfer_bar} shows \ours consistently outperforms human and Automatic Prompt Engineer (APE)~\citep{zhou2022large} baselines across \model{GPT-3.5}, \model{GPT-4}, and \model{PaLM 2}, yielding improvements by 9.1\%, 7.7\% and 6\% over APE, respectively. Extensive qualitative results further highlight the expert-level aspects of optimized prompts, indicating that \ours effectively bridges the domain gap in challenging tasks, offering great exploration efficiency and generalizability. As we anticipate the emergence of even more powerful LLMs that can understand intricate instructions, we believe that expert-level prompting will spearhead the next era of prompt engineering, where \ours stands as a pioneering step in this research direction.

\vspace{-10pt}
\section{Related Works}
\vspace{-10pt}

\noindent \textbf{Prompt optimization}.
Automatically discovering optimal prompts has emerged as a central challenge in the era of LLMs. For open-sourced LLMs, one can leverage their internal states or gradients to either train additional parameters, such as soft prompts~\citep{li-liang-2021-prefix, lester2021power, hu2021lora, wang2022multitask}, or search for discrete prompts via gradient-based search~\citep{shin2020autoprompt, wen2023hard} or reinforcement learning~\citep{deng2022rlprompt, zhang2022tempera}. However, such methods are less feasible for closed-sourced LLMs, which urges people to study gradient-free prompt optimization, typically assuming only APIs and a limited training set are available. Most gradient-free methods follow an iterative process of prompt sampling, i.e., starting from an initial prompt, they iteratively sample prompt candidates and score them to select the best one for the next iteration. Numerous methods emphasize diversifying the prompt candidates---examples include edit-based methods like deleting or swapping phrases~\citep{prasad2023grips}, back translation~\citep{xu2022gps}, evolutionary operations~\citep{guo2023connecting, fernando2023promptbreeder}, or more relevantly, LLM rewriting based on natural language feedback~\citep{zhou2022large, pryzant2023automatic, yang2023large}. There are also explorations into alternate sampling procedures like Monte Carlo search~\citep{zhou2022large}, Gibbs sampling~\citep{xu2023reprompting} or Beam search~\citep{pryzant2023automatic}. Nevertheless, \ours fundamentally differs from all the above methods in two ways. First, while primary search algorithms have been investigated~\citep{zhou2022large, xu2023reprompting, pryzant2023automatic}, we are the first to introduce strategic planning into prompting optimization research. This innovation provides a structured way to efficiently navigate the intricate space of prompts, with principled capabilities like lookahead and backtrack. Second, most previous methods generate prompt candidates as local variants, such as paraphrasing or LLM sampling, fail to incorporate fine-grained domain insights. Instead, we formulate prompt generation as the state transition and strategically convert error feedback into new states, leading to expert-level prompts.

\noindent \textbf{Augmenting LLMs with self-reflection and planning}.
Despite their remarkable capabilities, modern LLMs exhibit certain limitations, such as long-term coherence~\citep{malkin2022coherence}, lacking an internal world model~\citep{hao2023reasoning}, the inability to act in the real world, etc. Thus, augmenting LLMs with external modules like reasoning and tools has drawn extensive attention recently~\citep{mialon2023augmented, ozturkler2022thinksum, hao2023toolkengpt, jojic2023gpt}, of which two common strategies are relevant here: self-reflection and planning with LLMs. \textit{Self-reflection} encourages the LLM to introspect, critique its outputs, and subsequently suggest more refined solutions~\citep{jang2023reflection, pan2023automatically}. This has been leveraged to enhance a variety of applications, from complex computer tasks~\citep{shinn2023reflexion}, text generation~\citep{welleck2022generating} to reasoning~\citep{paul2023refiner}. 

Moreover, \textit{planning with LLMs} sheds light on evaluating and enhancing these models. At its core, planning is an essential ability for intelligent agents to generate a sequence of actions in achieving specific goals~\citep{mccarthy1963situations, bylander1994computational}. One line of research is to prompt and evaluate LLMs on planning tasks directly~\citep{liu2023llm+}. For instance, translation-based approaches translate natural language instructions into executable programs (e.g., Planning domain description language) to run classical planning algorithms. Another closer line of research is to augment the strategic reasoning ability of LLMs with planning-based algorithms. For example, Tree of Thoughts (ToT) applies DFS/BFS to augment CoT prompting, while both CoRe~\citep{zhu2022solving} and RAP~\citep{hao2023reasoning} utilize MCTS to navigate richer reasoning paths. Yet, in contrast to existing endeavors in LLM augmentation, \ours is the first novel framework for synergistically marrying the spirits of self-reflection and planning specifically tailored for prompt optimization.

\vspace{-8pt}
\section{Methodology}
\label{sec:method}
\vspace{-5pt}

Given a base LLM $\mathcal{B}$ and a target task $\mathcal{T}$, the job at hand for a prompt engineer is to craft an optimized natural language prompt $\mathcal{P}^{\mathcal{T}}$ that maximizes the performance of $\mathcal{B}$ on $\mathcal{T}$. However, the gap between novice and expert prompt engineers can be significant, particularly for tasks demanding specialized domain expertise, such as in the biomedical domain. Our primary objective is to autonomously refine the task prompt $\mathcal{P}^{\mathcal{T}}$ to bridge this knowledge gap, minimizing human intervention. Most existing approaches rely on sampling local prompt alternatives iteratively, which is not only resource-intensive but also lacks assurance of yielding an optimal final prompt. In light of this, we introduce \ours, an agent-based framework to produce expert-level task prompts via strategic planning and reflecting with error feedback during the prompting process, striking a proper balance of exploration and performance.

\begin{figure}
    \centering
    \includegraphics[width=0.9\linewidth]{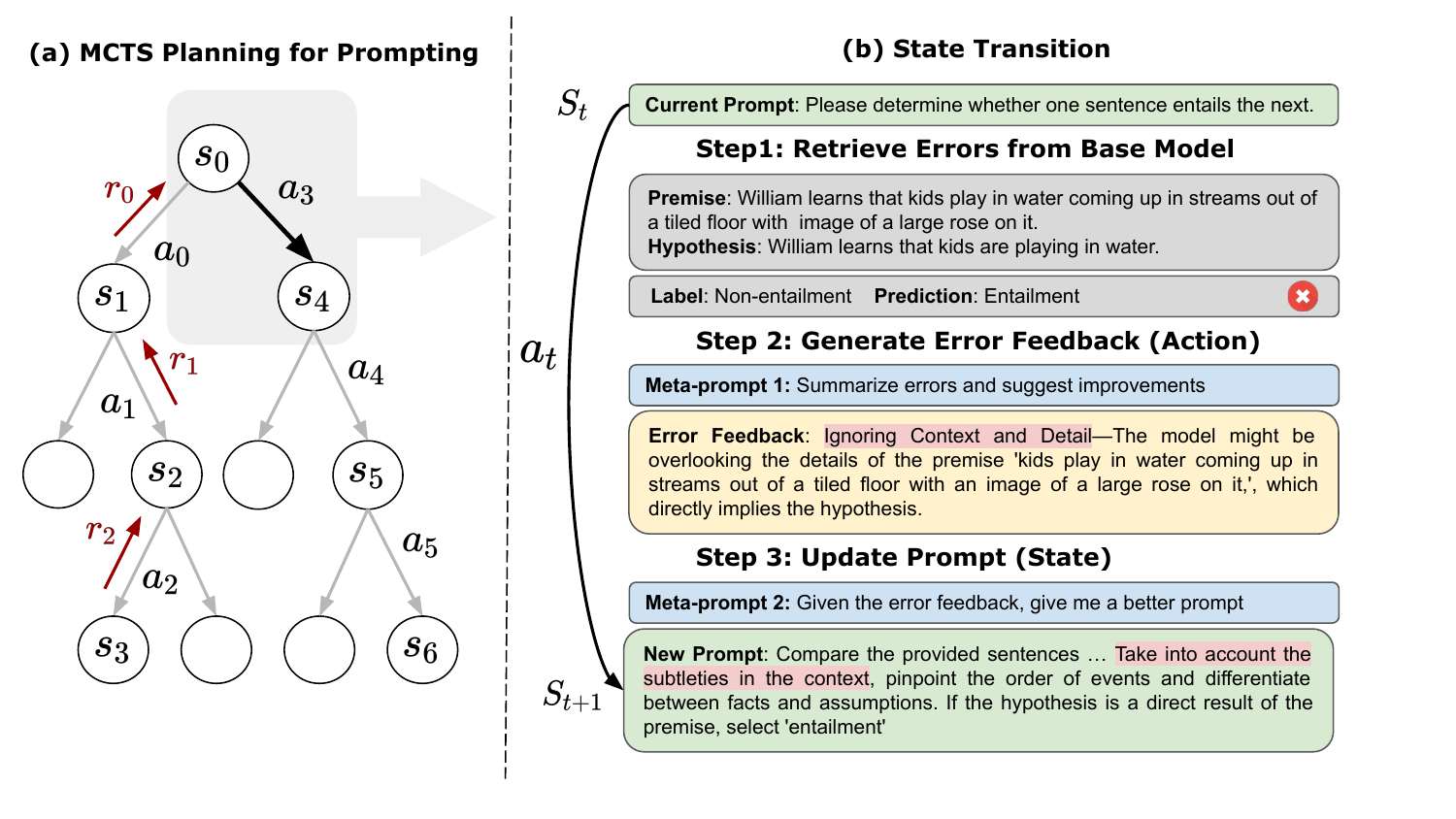}
    \vspace{-5pt}
    \caption{
    (a) MCTS (Monte Carlo Tree Search) planning for expert-level prompting. The tree structure enables strategic planning for \ours.
    (b) A simplified state transition example. Given a current state (prompt), the base model (\model{gpt-3.5-turbo}) collects errors from the task dataset. The optimizer model (\model{gpt-4}) provides error feedback accordingly. The optimized model then updates the prompt according to the feedback and transits to the next state.}
    \vspace{-15pt}
    \label{fig:framework}
\end{figure}

\noindent \textbf{Problem formulation.} Following a standard setting in prompt optimization~\citep{zhou2022large}, we start with an initial natural language task prompt $\mathcal{P}_0$ (e.g., ``Let's solve this problem step-by-step'') and a small set of training samples from target task $\mathcal{T}$ as $(Q, A)=\{q_i, a_i\}_{i=1}^N$, where $q_i$/$a_i$ are input/output pairs for each sample (e.g., a question and its answer). Given the model input consisting of $\mathcal{P}$ and $q_i$, the base LLM $\mathcal{B}$ makes the prediction (typically through a left-to-right generation process) based on $p_{\mathcal{B}}(a_i|q_i, \mathcal{P})$\footnote{Note this is traditionally a zero-shot setting we focus on, where task prompt excludes any training samples.}. The goal of prompt optimization is to find the optimal natural language prompt $\mathcal{P}^*$ that maximizes the performance towards a measure function $\mathcal{R}$ (e.g., accuracy). This can be formally defined as an optimization problem: 
$\mathcal{P}^* = \argmax_{\mathcal{P}\in \mathcal{S}}\ \sum_{i} \mathcal{R} (p_{\mathcal{B}}(a_i|q_i, \mathcal{P}))$,
where $\mathcal{S}$ denotes the sample space for a natural language prompt, an infinite and intractable space, if not impossible, to comprehensively enumerate. Conventionally, human experts draw upon a blend of heuristics and domain-specific insights to craft such prompts. Although previous optimization methods have attempted to leverage iterative sampling methods for prompt discovery~\citep{zhou2022large}, we advance this line of research by proposing a unified framework that seamlessly integrates strategic planning for superior, expert-level prompt optimization. Next, we introduce the formulation of \ours and then present the planning-based prompt optimization.

\vspace{-5pt}
\subsection{PromptAgent Framework Design}
\label{sec:framework}
\vspace{-5pt}

The goal of \ours is to effectively integrate expert prior knowledge into the task prompt while ensuring an efficient and strategic exploration of the expansive prompt space. In this planning framework, we define the state as each iteration or version of the task prompt, $s_t = \mathcal{P}_t$. This allows systematic monitoring of the evolution of prompts and directly applying refinements to modify them. Actions, in this context, can be thought of as potential modifications to the current prompt (state), such as word replacements or paraphrasing, as explored in prior works~\citep{jiang2020can, prasad2023grips}. However, a more desirable action space should introduce more effective and meaningful revisions that invoke prior expert knowledge, ultimately steering toward expert-level prompts. We thus propose error-based actions where each action is generated based on certain errors made by the base model. Specifically, as illustrated in Figure~\ref{fig:framework} (b), actions are framed as error feedbacks to guide subsequent refinements of the prompt. Such error feedbacks effectively suggest potential directions for correcting model errors, ensuring the revised prompt better instructs the base model to avoid previously observed pitfalls. Note that this approach also resonates with recent findings on the self-reflection capabilities of LLMs~\citep{pryzant2023automatic, shinn2023reflexion, paul2023refiner}, such that an LLM can directly reflect on their errors to yield better prompt modifications.

Given the definition of state and action, \ours formulates the prompt optimization problem as a Markov Decision Process (MDP) by the tuple $(\mathcal{S}, \mathcal{A}, {T}, r)$. Here, $\mathcal{S}$ denotes the state space, $\mathcal{A}$ is the action space, $T$ defines the transition function $T: \mathcal{S}\times \mathcal{A}\mapsto \mathcal{S}$, and $r$ is the reward function $r: \mathcal{S} \times \mathcal{A} \mapsto \mathbb{R}$. As illustrated in Figure~\ref{fig:framework} (a), for any given current state $s_t$, \ours iteratively generates an action $a_t$ based on $a_t \sim p_{\mathcal{O}}(a | s_t, m_1)$, where $m_1$ is a meta-prompt employed by an optimizer LLM $\mathcal{O}$ to facilitate the action generation. Specifically, Figure~\ref{fig:framework} (b) shows the two-step process of action generation: collecting errors of the base model from training samples (Step 1) and reflecting on such errors to draw useful error feedbacks (Step 2). Afterward, \ours obtains a new state based on the transition function $p_{\mathcal{O}}(s_{t+1} | s_t, a_t, m_2)$, where $m_2$ is another meta-prompt helping the state transition to update the prompt, also operating on $\mathcal{O}$. More specifically, given current error feedback as action $a_t$, $m_2$ asks the optimizer to generate a new prompt (state) to leverage any domain knowledge and effectively address model errors, similar to how prompting experts revise their prompts based on error feedbacks.

Finally, the quality of each newly generated state $s_t$ after applying action $a_t$ is determined by the reward function $r_t = r(s_t, a_t)$. Drawing parallels with the intricate nature of reward engineering in Reinforcement Learning (RL), crafting rewards could be complex to accommodate domain-specific knowledge or preferences specified for the task of interest. Without losing the generality of our framework across a variety of tasks, we straightforwardly define the reward as the task performance on a held-out set separated from the given training samples. The exact definition of reward, however, will depend on task-specific metrics as described in the implementation details later.

\vspace{-5pt}
\subsection{Strategic Planning for Prompt Optimization}
\label{sec:planning}
\vspace{-5pt}

The aforementioned reformulation of the prompt optimization enables us to seamlessly integrate \ours with principle planning algorithms, notably the Monte Carlo Tree Search (MCTS). This enables strategically navigating the vast prompt space while balancing the exploration and exploitation in finding high-reward paths of error feedbacks, which leads to the most generalizable expert-level prompts. Specifically, we observe some error feedbacks (actions) may inject instance-specific details into task prompts (states) that are hard to generalize task-wise (exploitation), where we need strategic planning to explore novel error feedbacks for higher rewards (exploration). MCTS operationalizes such strategic planning, as shown in Figure~\ref{fig:framework} (a), by progressively constructing a tree structure with each node as a state and each edge as the action for transiting states. MCTS expands the tree strategically by maintaining a state-action value function, $Q : \mathcal S \times \mathcal A \mapsto \mathbb R$, which represents the potential future rewards for applying an action $a_t$ to a state $s_t$. In other words, we rely on this function, ${Q} (s_t, a_t)$, to look ahead and estimate the potential rewards for paths following the current state-action pair. To update this $Q$ function and expand the tree, MCTS iteratively performs four operations: \textit{selection}, \textit{expansion}, \textit{simulation}, and \textit{back-propagation}. The iteration process ends when a pre-defined number of iterations is reached, and we then select the highest-reward trace for the final prompt. We next explain the four operations in \ours, and the pseudocode of our MCTS-based prompt optimization can be found in Algorithm~\ref{alg:mcts} of the Appendix.

\noindent \textbf{Selection} is the first step that selects the most promising nodes at each level to be further expanded and explored. At each iteration, it starts from the root node $s_0$, traverses through each tree level, selects a subsequent child node at every level, and stops at a leaf node. When selecting the child node at each level, we leverage the \textit{Upper Confidence bounds applied to Trees} (UCT) algorithm, which is well-known for balancing the exploitation (choosing high-value nodes) and exploration (choosing less-visited nodes) as follows:
\begin{align}
\label{eq:uct}
\vspace{-5pt}
    a^\ast_t = \argmax_{a_t' \in {A}(s_t)} \left( Q(s_t, a_t') + c \cdot \sqrt{\frac{\ln \mathcal{N}(s_t)}{\mathcal{N}(\text{ch}(s_t, a_t'))}}\ \right)
\vspace{-5pt}
\end{align}
\noindent where ${A}(s_t)$ is the action set for node $s_t$, $\mathcal{N}(s_t)$ is the number of visiting times for node $s_t$, $\text{ch}(s, a)$ represents the child node for $s_t$ after applying action $a_t'$ and $c$ is a constant to adjust the exploration. As we can see, the first term signifies exploitation by the $Q$ value, and the second term indicates exploration, measuring the uncertainty for less visited nodes. In other words, if a node was less explored and its child node was less visited before, the second term will be higher. 

\noindent \textbf{Expansion} grows the tree by adding new child nodes to the leaf node reached by the previous \textit{selection} step. This is done by applying the action generation and state transition (Figure~\ref{fig:framework} (b)) multiple times, resulting in multiple new actions and states. Note that we may sample multiple training batches to derive diverse error feedbacks (actions). Within new nodes, we then send the highest-reward one to the next \textit{simulation} step.

\noindent \textbf{Simulation} is the lookahead step to simulate the future trajectories for the selected node from the previous \textit{expansion} step. This step usually comes with a playout policy to reach the terminal state quickly and calculate the future rewards. The choice of playout could be flexible, such as choosing random moves until the terminal. To reduce the computation cost of simulation and simplify the process, we perform the previous \textit{expansion} step iteratively until the terminal, i.e., we keep generating multiple actions and selecting the highest-reward node among them to proceed to the next tree level.

\noindent \textbf{Back-propagation} happens when a terminal state is met during the \textit{simulation}. The terminal state is usually defined when a pre-defined maximum depth is reached, or an early-stopping criterion is encountered. We then back-propagate the future rewards along the path from the root to the terminal node by updating the $Q$ value function. Specifically, for each state-action pair in the path, $Q(s_t, a_t)$ is updated by aggregating the rewards from all future trajectories starting from $s_t$ as follows:
\begin{align}
\vspace{-5pt}
\label{eq:backpropagate}
Q^*(s_t, a_t) = \frac{1}{M} \sum_{j=1}^{M} \left( \sum_{s' \in S_{s_t}^j, a' \in A_{a_t}^j} r(s', a') \right)
\vspace{-5pt}
\end{align}
\noindent where $M$ is the number of future trajectories starting from $s_t$, $S_{s_t}^j$ and $A_{a_t}^j$ represent the $j$-th state and action sequences starting from $s_t$ and $a_t$, respectively.

\ours executes the above four operations with a pre-defined number of iterations to stabilize the $Q$ values and fully grow the tree for exploring the vast prompt space. We finally need to select the best trace and node (i.e., prompt) for the final evaluation. Multiple alternative solutions can be leveraged for this output strategy, e.g., one could opt for the best node in the best path with the highest reward, or directly choose the leaf node with the largest number of visiting times. For simplicity and empirical purposes, we use the first strategy to select the output prompt, which works the best in our experiments.

\vspace{-5pt}
\section{Experiments}
\vspace{-5pt}

\subsection{Experimental Setup}
\vspace{-5pt}

\noindent \textbf{Tasks and Datasets.}\label{dataset}
To comprehensively evaluate the effects of expert-level prompt optimization for a wide range of applications, we curate 12 tasks from three distinct domains for thorough experiments: \textit{BIG-Bench Hard (BBH)}, as well as \textit{domain-specific} and \textit{general NLP} tasks. BBH~\citep{suzgun2022challenging} is a subset of challenging BIG-Bench tasks~\citep{srivastava2023beyond} that are beyond the capabilities of current LLMs. We select 6 BBH tasks that emphasize a blend of domain knowledge (i.e., Geometric Shapes and Causal Judgment) and complex reasoning abilities (i.e., Penguins in a table, Object Counting, Epistemic Reasoning, and Temporal Sequences). We also select three domain-specific tasks in the biomedical domain, where domain insights are explicitly desired when crafting expert-level prompts. Such tasks include a disease named-entity recognition (NER) task (NCBI, \citet{dougan2014ncbi}), a biomedical sentence similarly task (Biosses, \citet{souganciouglu2017biosses}), and a medical question answering task (Med QA, \citet{jin2021disease}). Moreover, to show \ours can also be generally applicable and beneficial for traditional NLP tasks, we further select three well-known NLU tasks, i.e., two text classification tasks (TREC, \citet{voorhees2000building} and Subj, \citet{pang2004sentimental}), and a natural language inference task (CB, \citet{de2019commitmentbank}).

\noindent \textbf{Baselines.}
We compare our methods with three types of baselines: ordinary human prompts, Chain-of-Thought (CoT) prompts, and recent prompt optimization methods. (1) \textit{Human prompts} are human-designed instructions representing the generic level of prompt engineering, which usually come from the original datasets. We also have a few-shot (FS) version of human prompts with teaching examples from \cite{suzgun2022challenging} for BBH tasks and randomly sampled ones from the training set for others. (2) \textit{CoT prompts} are considered very effective tricks to boost LLM performance by inducing intermediate reasoning steps, especially for BBH tasks~\citep{suzgun2022challenging}. We directly use the CoT prompts from \cite{suzgun2022challenging} for BBH tasks and construct CoT prompts by ourselves for other tasks. We also have a zero-shot (ZS) version of CoT, using ``Let's think step by step'' as the prompt to trigger CoT behavior without few-shot examples~\citep{kojima2022large}. (3) Prompt optimization methods include \textit{GPT Agent} and \textit{Automatic Prompt Engineer (APE)}~\citep{zhou2022large}. {GPT Agent} represents the recent surge of interest in LLM-powered autonomous agents~\citep{weng2023prompt}, such as Auto-GPT\footnote{\url{https://github.com/Significant-Gravitas/AutoGPT}}. Such agents are expected to autonomously perform planning and self-reflection to solve human requests, including optimizing task prompts. We leverage one of the powerful ChatGPT Plugins~\citep{ChatGPT_plugins} with GPT-4, \textit{AI Agents}\footnote{\url{https://aiagentslab.com/}} for prompt optimization. Specifically, similar to \ours, we sample similar model errors and ask \textit{AI Agents} plugin to rewrite the prompt based on the errors with a similar iteration number as \ours. Lastly, APE is one of the most recent prompt optimization methods that proposes a Monte Carlo search-based method to iteratively propose and select prompts. 

\begin{table}[!t]
\caption{ Prompting performance on BBH tasks. ZS: Zero-Shot, FS: Few-Shot. 
We select six challenging tasks from BBH~\citep{suzgun2022challenging}, requiring domain knowledge (e.g., Geometry) or reasoning (e.g., Causal Judgement).
Our method outperforms in 5/6 tasks, with only CoT surpassing in Object Counting. On average, our accuracy exceeds others by at least 9\%.}
\label{tab:bigbench} 
\vspace{-5pt}
\definecolor{Gray}{gray}{0.90}
\newcolumntype{a}{>{\columncolor{Gray}}c}
\centering
\resizebox{0.98\linewidth}{!}{%
\begin{tabular}{@{}lcccccca@{}}
\toprule
              & Penguins & Geometry & Epistemic & Object Count. & Temporal & Causal Judge. & \textit{Avg.} \\ \midrule
Human (ZS)     &     0.595      &     0.227     &      0.452     &        0.612          &    0.720      &        0.470           &     0.513 \\
Human (FS)      &     0.595      &     0.315     &     0.556      &         0.534         &     0.408     &        0.620           &    0.505  \\
CoT (ZS) &     0.747      &     0.320     &     0.532      &         0.542         &    0.734      &        0.610           &    0.581  \\
CoT          &     0.747      &     0.540     &     0.720      &        \textbf{0.960}          &   0.626      &        0.650           &    0.707  \\
GPT Agent     &     0.696      &     0.445     &     0.406      &         0.502         &    0.794      &        0.520           &   0.561   \\
APE           &     0.797      &     0.490     &     0.708      &       0.716           &      0.856    &        0.570           &   0.690   \\
PromptAgent   &    \textbf{0.873}      &     \textbf{0.670}     &     \textbf{0.806}      &       \textbf{ }0.860\textbf{ }         &    \textbf{0.934}      &        \textbf{ 0.670 }         &      \textbf{0.802}\\ \bottomrule
\end{tabular}
}
\vspace{-15pt}
\end{table}

\noindent \textbf{Implementation details.} For the datasets with default testing or validation set, we use their original split to obtain our testing set. If there is no official training/testing split, such as BBH tasks, we sample a reasonably large set for stable testing. As stated in Section~\ref{sec:framework}, we also split a portion of training samples for calculating the reward. The details of the datasets can be found in Appendix~\ref{input_and_dataset}. 
Unless further specified, we select \model{GPT-3.5} as the default base LLM to be optimized, which is one of the decently powerful modern LLMs. For the optimizer LLM, we need one with a good self-reflection ability and, thus, use \model{GPT-4} as the default optimizer LLM. We set the temperature as 0.0 for base LLM to make predictions and 1.0 in other contexts. 
When implementing \ours, we set the number of iterations for MCTS as 12, and the exploration weight $c$ in Equation~\ref{eq:uct} as 2.5. During the expansion step, we generate actions based on model errors by sampling batches from training samples. We sample \textit{expand\_width} batches and generate \textit{num\_samples} new prompts per batch. The maximum depth of each path is \textit{depth\_limit}. To simplify the process of tuning these hyperparameters, we explore three settings: 
\textit{Standard}, \textit{Wide}, and \textit{Lite}, where \textit{Standard} and \textit{Lite} have larger depth, while \textit{Wide} generates more nodes per expansion step  (Specific parameters can be found in Appendix Table~\ref{tab:hyperparameter}). The best setting for \ours is selected based on the rewards. Further details are available in Appendix~\ref{sec: experimental setting}, including input formatting, data splitting, and the implementation specifics of both the PromptAgent and baseline methods.

\vspace{-5pt}
\subsection{Results and Analyses}
\vspace{-5pt}

\begin{table}[!t]
\caption{{Prompt performance on specialized and general NLU tasks. Specialized tasks are three biomedical tasks explicitly asking for domain knowledge for prompting. General NLU tasks are used to demonstrate the generality of our method. Ours significantly outperformed in all tasks.}}
\label{tab:other_tasks} 
\vspace{-5pt}
\definecolor{Gray}{gray}{0.90}
\newcolumntype{a}{>{\columncolor{Gray}}c}
\centering
\resizebox{0.95\linewidth}{!}{%
\begin{tabular}{@{}lcccaccca@{}}
\toprule
              & \multicolumn{4}{c}{Domain-specific Tasks} & \multicolumn{4}{c}{General NLU Tasks} \\ \cmidrule(lr){2-5} \cmidrule(lr){6-9}
              & NCBI (F1)  & Biosses  & Med QA  & \textit{Avg.} & Subj   & TREC   & CB   & \textit{Avg.}  \\ \midrule
Human (ZS)     & 0.521 &    0.550      &    0.508    &  0.526 &   0.517   &   0.742     &    0.714    &  0.658 \\
Human (FS)      & 0.447 &     0.625     &   0.492     &0.521 &  0.740    &    0.742    &   0.429     &  0.637 \\
CoT (ZS)        & 0.384 &     0.425     &    0.508     &0.439&  0.656    &    0.63    &   0.750     &           0.679 \\
CoT          &   0.376    &   0.675       &   0.542      &0.531&   0.670   &   0.784     &    0.643    &          0.699  \\
GPT Agent     &   0.125    &     0.625     &     0.468   & 0.406&   0.554   &    0.736    &    0.339    &           0.543 \\
APE           &   0.576    &     0.700     &    0.470    & 0.582&  0.696    &   0.834     &    0.804    &           0.778 \\
PromptAgent   &   \textbf{0.645}    &    \textbf{0.750}      &    \textbf{0.570}    & \textbf{0.655}&   \textbf{0.806}   &   \textbf{0.886}     &    \textbf{0.911}    &    \textbf{0.868}  \\
\bottomrule
\end{tabular}
}
\vspace{-15pt}
\end{table}

\noindent \textbf{Comparison with various prompting baselines.}
Table~\ref{tab:bigbench} \& \ref{tab:other_tasks} present a comprehensive comparison of expert-level prompts generated by \ours against human prompts, CoT prompts, and existing prompt optimization methods across 12 tasks spanning three domains. Observing BBH tasks from Table~\ref{tab:bigbench}, \ours significantly outperforms all baselines overall and achieves 28.9\%, 9.5\%, and 11.2\% relative improvement over baselines, i.e., human prompts (ZS), CoT, and APE, respectively. It is noteworthy that CoT prompts are especially effective in BBH tasks than human prompts, similar to findings from \cite{suzgun2022challenging}. This is because BBH tasks usually require strictly formatted solutions that can be readily induced by the step-by-step CoT reasoning, which also explains why CoT achieves very good performance on \textit{Object Counting} that can benefit from step-by-step solutions the most.
However, \ours still outperforms CoT by a great margin in all tasks (except \textit{Object Counting}), indicating that our optimized expert-level prompt can lead to bigger improvement over few-shot CoT reasoning (even under the zero-shot prompt setting). 
Regarding optimization methods, while we appreciate the planning and self-reflection of the GPT Agent, its planning is only used for a single turn of prompt rewriting, but not on a global scale of strategically exploring prompt space. APE, on the other hand, shows a greater scale of searching ability, but its exploration is based on Monte Carlo search, which suffers from inefficient planning and a lack of error-based reflections. Both deficits of GPT Agent and APE suggest the necessity of strategic planning in \ours to fully explore the prompt space and deliver expert-level prompts.

Table~\ref{tab:other_tasks} presents results on domain-specific and general NLP tasks. The former encompasses a broad spectrum of biomedical tasks, such as information extraction, sentence similarity, and question answering. Crafting prompts for these tasks requires extensive domain knowledge and heavy LLM prompt engineering instincts, where we can observe that straightforward human prompts and CoT prompts do not work very well. Prompt optimization methods like APE with automatic prompt sampling and refining are promising to incorporate domain knowledge without too much human intervention. Notably, \ours surpasses APE significantly by +7.3\% improvement on average, suggesting \ours can better induce effective domain knowledge to produce expert-level prompts and close the knowledge gap between novice and expert prompt engineers. For general NLP tasks, the efficacy and generality of \ours are further emphasized, outperforming both CoT and APE by margins of +16.9\% and +9\%, respectively. This implies the nontrivial expert gap, even for general NLP tasks, underscoring the imperative for expert prompts in diverse applications.

\noindent \textbf{Prompt generalization.}
We next conduct experiments to investigate whether our optimized prompts can be generalized to other base LLMs. This emphasizes the robustness and transferability of expert-level prompts, which are urgently favorable and underpinning two key facts: (a) the domain insights and nuanced guidance in expert prompts can be seamlessly transferred across powerful LLMs, reinforcing the universal applicability of expert prompts, and (b) we only need to optimize each task once, leading to better computational efficiency. It is crucial to note that the primary goal of \ours is to optimize prompts for state-of-the-art LLMs to achieve expert-level prompting, while less advanced and smaller LLMs, like GPT-2 or LLaMA, may not adeptly grasp the subtleties of these expert-level prompts, potentially causing significant performance drop. 
Nonetheless, for a holistic assessment, we evaluate two additional base LLMs, one more potent (\model{GPT-4}) and one less robust (\model{PaLM 2}) than \model{GPT-3.5}, within this experimental framework.

Table~\ref{tab:prompt_transfer} shows the results when we directly apply the optimized prompts from \model{GPT-3.5} to \model{GPT-4} and \model{PaLM 2} (\model{chat-bison-001}) across all 12 tasks. For comparison, we also adopt the same human and APE prompts to these base LLMs as baselines. For certain tasks, such as \textit{Penguins}, we may employ slightly different prompts than those referenced in Table~\ref{tab:bigbench} to make \model{PaLM 2} generate reasonable responses instead of persistent \textit{null} answers. 
Observing Table~\ref{tab:prompt_transfer}, it is worth highlighting that when a stronger base LLM as \model{GPT-4} is deployed, our expert prompts manifest further enhancements, either on par with or outperforming Human and APE prompts in almost all tasks (11/12) (The only exception, \textit{Temporal}, seems to be a solved task by \model{GPT-4} with almost perfect accuracy). This underscores the untapped potential of expert prompting, especially with the evolution of more sophisticated LLMs in the near future. When transferring expert prompts to a weaker LLM as \model{PaLM 2}, its performance drops dramatically across all tasks unexpectedly. Nonetheless, we still observe \ours exceeds both baselines on 7/12 tasks, with great improvements on domain-specialized tasks, such as \textit{NCBI}, demonstrating the usefulness of domain insights from expert prompts.

\begin{table}[!t]
\vspace{-2mm}
\caption{Prompt generalization results.
While we optimize \model{GPT-3.5} as the default base LLM, its optimized prompts are transferable to other base LLMs like \model{GPT-4} and \model{PaLM 2} (\model{chat-bison-001}). \model{GPT-4} sees further enhancement with our prompts, beating baselines in 11/12 tasks. Weaker LLMs like \model{PaLM 2} may have challenges with our advanced prompts but still surpass baselines in 7/12 tasks. Overall, ours can significantly beat baselines with different base LLMs.
}
\label{tab:prompt_transfer} 
\vspace{-1mm}
\definecolor{Gray}{gray}{0.90}
\newcolumntype{a}{>{\columncolor{Gray}}c}
\centering
\resizebox{0.95\linewidth}{!}{%
\begin{tabular}{@{}lccccccccc@{}}
\toprule
        & \multicolumn{3}{c}{GPT-3.5} & \multicolumn{3}{c}{GPT-4} & \multicolumn{3}{c}{PaLM 2} \\ \cmidrule(lr){2-4} \cmidrule(lr){5-7} \cmidrule(lr){8-10}
       Tasks  & Human  & APE  & Ours  & Human  & APE  & Ours & Human  & APE  & Ours  \\ \midrule
Penguins     & 0.595 & 0.747 & \textbf{0.797} & 0.772 & 0.848 & \textbf{0.962} & 0.430 & 0.443 & \textbf{0.456} \\
Geometry     & 0.227 & 0.490 & \textbf{0.670} & 0.495 & 0.445 & \textbf{0.680} & 0.290 & 0.215 & \textbf{0.360} \\
Epistemic    & 0.452 & 0.708 & \textbf{0.806} & 0.734 &  \textbf{0.848} & \textbf{0.848} & 0.470 & 0.392 & \textbf{0.588} \\
Object Count.  & 0.612 & 0.716 & \textbf{0.860} & 0.830 & 0.852 & \textbf{0.888} & 0.290 & \textbf{0.378} & 0.320 \\
Temporal     & 0.720 & 0.856 & \textbf{0.934} & 0.980 & \textbf{0.992} & 0.982 & 0.540 & 0.522 & \textbf{0.620} \\
Causal Judge. & 0.470 & 0.570 & \textbf{0.670} & 0.740 & 0.740 & \textbf{0.770} & \textbf{0.440} & \textbf{0.440} & 0.430 \\ \midrule
NCBI (F1)    & 0.521 & 0.576 & \textbf{0.645} & 0.588 & 0.428 & \textbf{0.697} & 0.016 & 0.025 & \textbf{0.177} \\
Biosses      & 0.550 & 0.700 & \textbf{0.750} & 0.700 & 0.775 & \textbf{0.800} & 0.500 & 0.300 & \textbf{0.600} \\
Med QA       & 0.508 & 0.470 & \textbf{0.570} & 0.770 & 0.758 & \textbf{0.774} & \textbf{0.284} & 0.274 & 0.276 \\ \midrule
Subj        & 0.517 & 0.696 & \textbf{0.806} & 0.867 & 0.805 & \textbf{0.879} & 0.496 & \textbf{0.537} & 0.499 \\
TREC         & 0.742 & 0.834 & \textbf{0.886} & 0.716 & 0.764 & \textbf{0.876} & 0.380 & \textbf{0.400} & 0.230 \\
CB           & 0.714 & 0.804 & \textbf{0.914} & \textbf{0.911} & 0.893 & \textbf{0.911} & 0.571 & 0.643 & \textbf{0.732} \\ \midrule
\textit{Average}          & 0.552 & 0.685 & \textbf{0.776} & 0.759 & 0.762 & \textbf{0.839} & 0.392 & 0.381 & \textbf{0.441} \\ \bottomrule 
\end{tabular}
}
\vspace{-15pt}
\end{table}

\begin{wraptable} {r}{0.5\linewidth}
\vspace{-12pt}
\caption{Ablation study on search methods. MC: Monte Carlo search, Greedy: greedy depth-first search, Beam: beam search. Testing tasks are representative of three domains from BBH~\citep{suzgun2022challenging}, domain-specialized and general NLU. Our method consistently outperforms all other ablated search algorithms across every task we evaluated.}
\label{tab:ablation} 
\vspace{-5pt}
\definecolor{Gray}{gray}{0.90}
\newcolumntype{a}{>{\columncolor{Gray}}c}
\centering
\resizebox{1\linewidth}{!}{%
\begin{tabular}{@{}lcccc@{}}
\toprule
          & MC    & Beam & Greedy  & MCTS (Ours) \\ \midrule
Penguins  & 0.772 & 0.823 & 0.810  & \textbf{0.873} \\
Biosses   & 0.575 & 0.675 & 0.700  & \textbf{0.750} \\
Geometry  & 0.490 & 0.610 & 0.545  & \textbf{0.670} \\
Causal    & 0.650 & 0.610 & 0.660  & \textbf{0.670} \\
Subj      & 0.692 & 0.765 & 0.778  & \textbf{0.806} \\ \midrule
\textit{Average} & 0.635 & 0.697 & 0.698  & \textbf{0.754} \\ \bottomrule
\end{tabular}
}
\vspace{-12pt}
\end{wraptable}

\noindent \textbf{Ablation on search strategies.}\label{sec:search_ablation}
To investigate the effect of strategic planning in \ours systematically, we conduct a thorough ablation study by comparing multiple alternative search strategies to MCTS, i.e., a single Monte Carlo (MC) search, a greedy depth-first search (Greedy), and a Beam search. We use the same action generation and state transition as in \ours and only replace the MCTS planning with each search method. Specifically, MC is a directionless search with a single step of randomly sampling and selecting one action. Greedy provides more structured exploration by consistently choosing the best among multiple samples per step. Beam search also focuses on a structured exploration by keeping multiple promising paths at each level. We keep the same number of overall explored prompts (exploration efficiency; see below for more results) for all three baselines to have a similar exploration space. See more implementation details about search variants in Appendix~\ref{search_baselines}. 

We select a subset of tasks from all three domains to compare all the above search variants due to the computation budget. Table~\ref{tab:ablation} shows that both Greedy and Beam greatly improve the MC baseline, suggesting the necessity of structured iterative exploration in our framework. When maintaining the same exploration efficiency, we observe comparable overall performance for Beam and Greedy. However, neither method strategically explores the prompt space since they operate in a strictly forward direction, lacking the capability to foresee future outcomes and backtrack to past decisions. In contrast, the strategic planning for MCTS allows \ours to navigate complex expert prompt spaces more effectively, which significantly surpasses all search ablations on all tasks and gets a relative 5.6\% overall improvement over the best baseline.

\begin{table}[t]
\caption{
Prompt comparison for the NCBI task, including normal human prompt, APE-optimized prompt, and expert-level prompt optimized by PromptAgent. Both baselines mostly describe the task, while our expert prompt is composed of more complex structures and domain-specific insights, achieving superior performance. {Bold text} denotes \textbf{domain knowledge} usually handcrafted by domain specialists, but here automatically discovered by \ours. We highlight different aspects of expert prompt with colors, including \ctext[RGB]{233,252,232}{Task Description}, \ctext[RGB]{255,230,230}{Term Clarification}, \ctext[RGB]{230,246,255}{Solution Guidance}, \ctext[RGB]{255,230,200}{Exception Handling}, \ctext[RGB]{255,225,255}{Priority \& Emphasis}, \ctext[RGB]{230, 230, 255}{Formatting}. (Best view with colors)
}
\label{tab:ncbi_highlights} 
\definecolor{Gray}{gray}{0.90}
\newcolumntype{a}{>{\columncolor{Gray}}c}
\centering
\resizebox{0.9\linewidth}{!}{%
\begin{tabular}{@{}lp{10.2cm}c@{}}
\toprule
Approach    & Optimized Prompt & F1 score. \\ \midrule
Human       &      \ctext[RGB]{233,252,232}{Extract the disease or condition from the sentence, if any is mentioned.}            &    0.521  \\\addlinespace
APE         &       \ctext[RGB]{233,252,232}{If any disease or condition is mentioned in the sentence, extract it.}        &     0.576 \\\addlinespace
PromptAgent &  
\ctext[RGB]{233,252,232}{You're tasked with extracting diseases or conditions from the given sentence, }\ctext[RGB]{255,230,200}{remember to be cautious and \textbf{avoid incorporating any associated elements such as inheritance patterns (like autosomal dominant), genes or gene loci (like PAH), proteins, or biological pathways}.}\ctext[RGB]{230,246,255}{ The task does not entail making assumptions or inferences about the disease names based on other advanced biological terms in the context. \textbf{Consider both specific diseases and broader categories, and remember diseases and conditions can also appear as common abbreviations or variations}. }\ctext[RGB]{230, 230, 255}{Provide the identified diseases or conditions in this format: \{entity\_1,entity\_2,....\}. If there are no diseases or conditions present, output an empty list in this form: \{\}. }\ctext[RGB]{255,230,200}{\textbf{Note that the term `locus' should be recognized as a genomic location and not a disease name}.}     & 0.645

\\ \bottomrule
\end{tabular}
}
\vspace{-10pt}
\end{table}

\begin{figure}[!t]
    \centering
    \begin{subfigure}[b]{0.55\textwidth}
        \centering
        \includegraphics[scale=0.53]{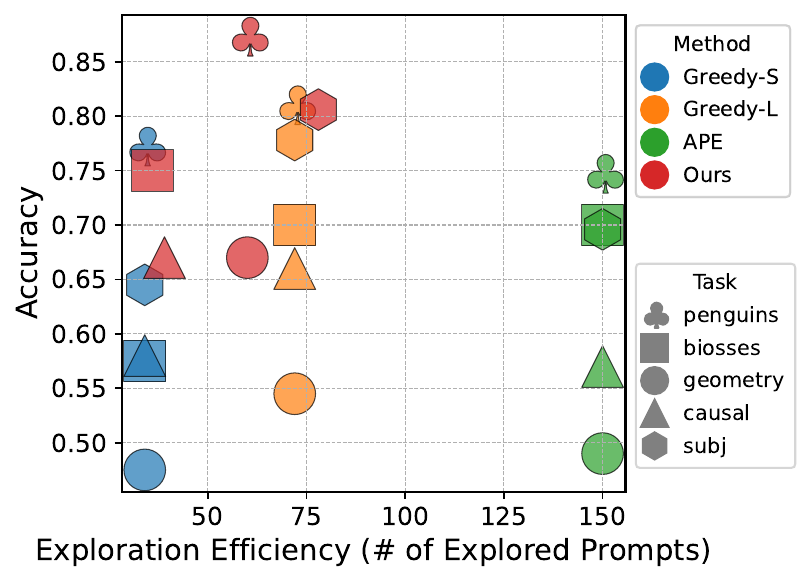}
        \vspace{-5pt}
        \caption{Performance vs. Exploration Efficiency}
        \label{fig:efficiency}
    \end{subfigure}
    \hfill
    \begin{subfigure}[b]{0.43\textwidth}
        \centering
        \includegraphics[scale=0.33]{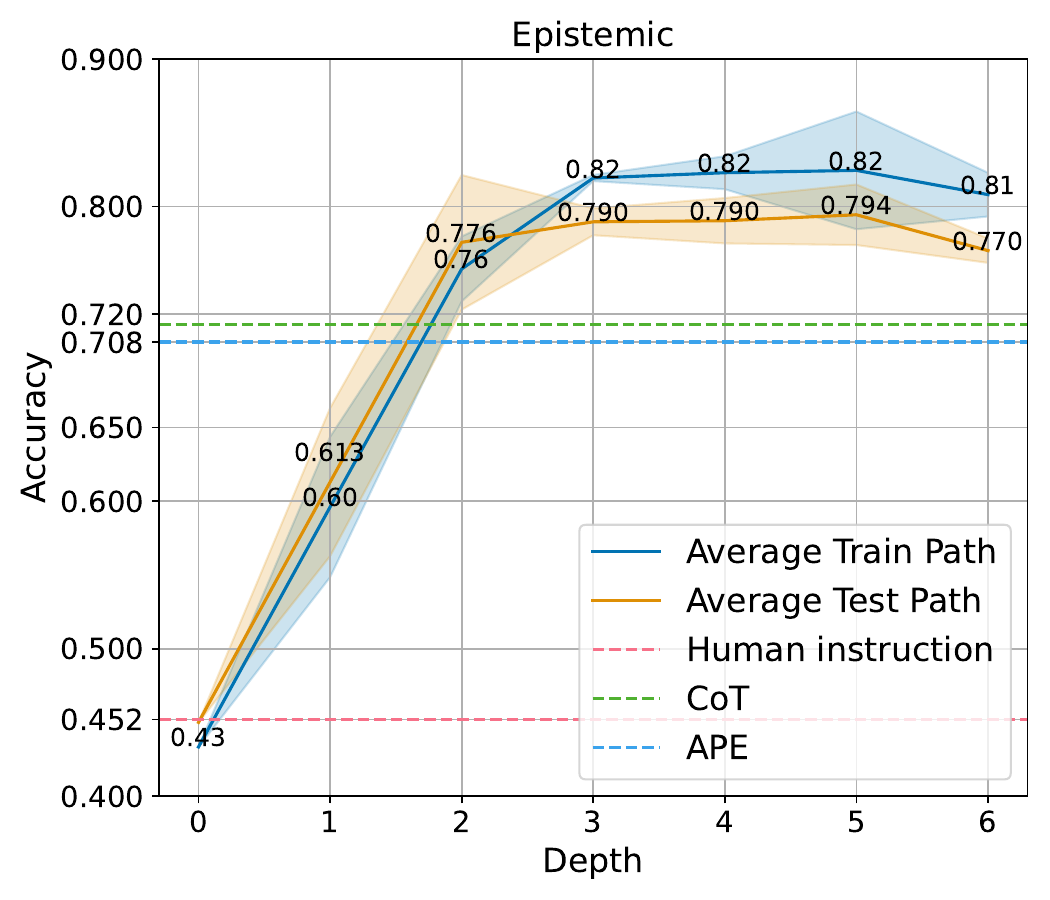}
        \vspace{-5pt}
        \caption{Convergence Analysis}
        \label{fig:convergence}
    \end{subfigure}
    \vspace{-5pt}
    \caption{
    (a) Exploration efficiency analysis. A proper balance of exploration and exploitation is crucial for search and planning. 
    We compare the number of explored prompts between our method and three strong baselines. Ours achieves the best trade-off of performance and exploration (clustering in the top-left corner).
    (b) Convergence curves for \textit{Epistemic} task. We visualize the mean and variance of the training and testing performance along the paths. We can observe that both curves increase at first and become stable after depth 3, suggesting a stable learning process}
    \label{fig:efficiency_convergence}
    \vspace{-10pt}
    \label{fig:f3}
\end{figure}

\noindent \textbf{Exploration efficiency analysis.}
In addition to the superior performance, one of the key advantages of \ours is that it can efficiently explore the prompt space via strategic planning. Exploration efficiency is also vital to make the computation cost of the search manageable. We thus analyze the exploration efficiency by comparing \ours with some of our search baselines, including Greedy Search and APE from the previous section. Specifically, the exploration efficiency is measured by the number of prompts explored during the search, i.e., nodes generated during the exploration. We plot its relationship with the task performance in Figure~\ref{fig:efficiency}. The Greedy-S and Greedy-L are based on Greedy Search with 34 and 72 explored prompts. The APE explores 150 prompts in each task. The figure shows that points of \ours are clustered around the top left corner, indicating a superior performance with higher accuracy but fewer explored nodes (higher exploration efficiency). Notably, while increasing the number of prompts in Greedy Search may enhance performance (from Greedy-S to Greedy-L), it demands higher exploration cost and still does not surpass \ours. Also, without principled guidance, directionless searches like APE cannot effectively boost performance, even with larger exploration. Nevertheless, to maintain exploration efficiency and superior performance, strategic planning is crucial in \ours and worthy of further research investment in future works. The detailed hyperparameter settings of Greedy-S, Greedy-L, and APE are in Appendix~\ref{search_baselines}

\noindent \textbf{Convergence analysis.}
To delve deeper into the learning process of \ours, we examine the evolution of expert prompts throughout the tree planning process. Specifically, we monitor and visualize performance changes with respect to tree depth. As illustrated in Figure~\ref{fig:convergence} for the \textit{Epistemic} task, we assess the performance across all nodes and aggregate both training (reward) and testing performance at each depth level. The plotted trajectories represent the evolution of average performance on both training (reward) and testing, illustrating a consistent improvement and gradually surpassing all baseline methods. For brevity, convergence plots for other tasks and hyperparameter settings, focusing solely on training trajectories to reduce computational overhead on testing sets, are provided in Appendix~\ref{sec: Convergence observation} and Appendix~\ref{sec: hyperparameter}. A recurring pattern observed, similar to that in Figure~\ref{fig:convergence}, indicates an upward trend in the initial iterations, suggesting a robust learning dynamic of \ours to iteratively refine and enhance expert prompts.

\begin{figure}
    \centering
    \includegraphics[width=0.9\linewidth]{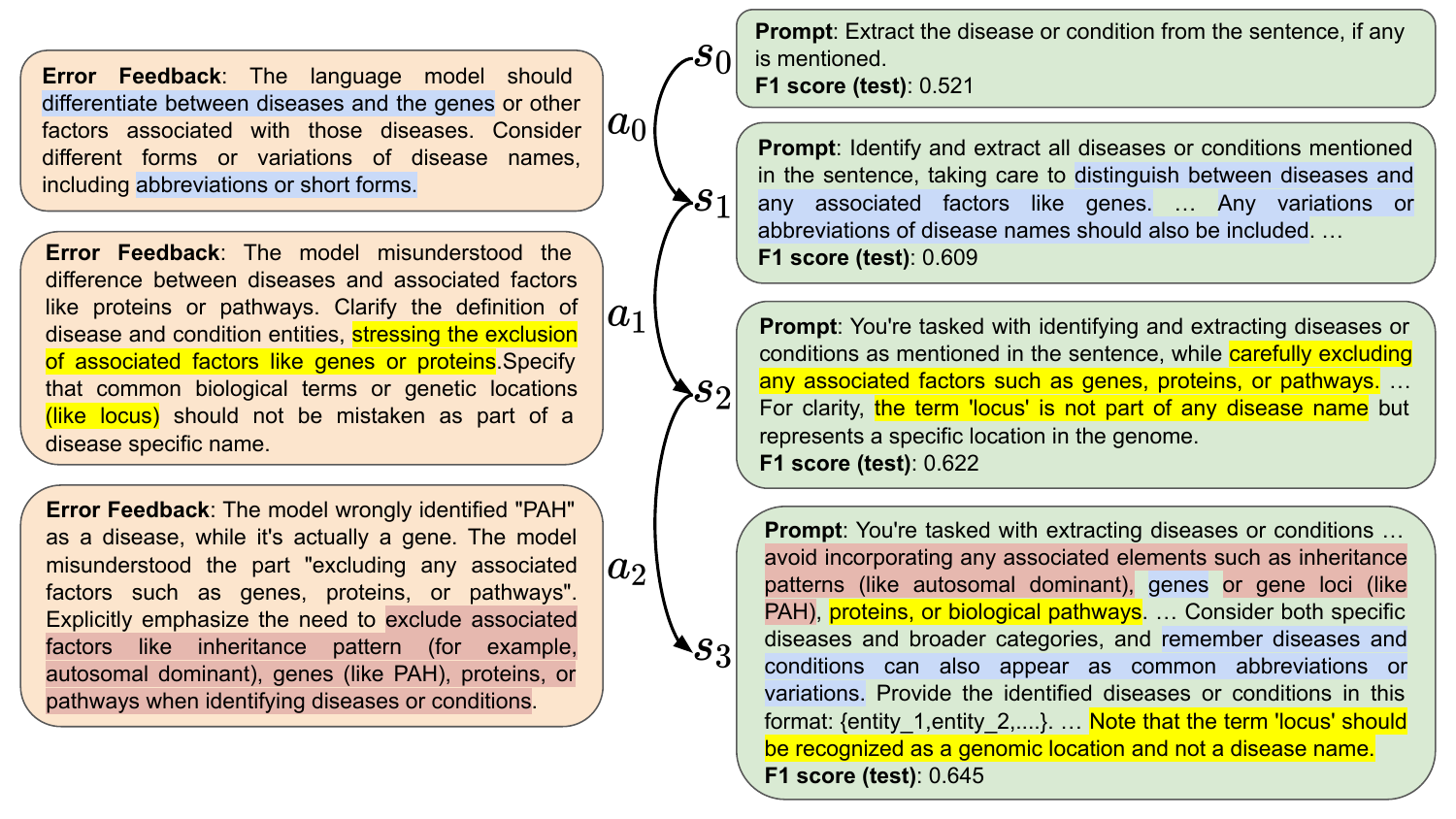}
    \vspace{-5pt}
    \caption{
    The MCTS state-action transition trajectory of the highest average reward path in NCBI. The initial state is $s_0$ with a human-written prompt. At each state transition step, a new prompt is crafted by adjusting the prior state based on error feedback. Highlighted colors indicate similar domain-specific insights. The last state integrates the information from the entire trajectory, elevating the F1 score from 0.521 to 0.645.}
    \label{fig:trace}
    \vspace{-15pt}
\end{figure}

\noindent \textbf{Qualitative analysis.}
To provide a more direct illustration of how \ours progressively leverages error feedback (action) to enhance prompts (states), we conduct a qualitative analysis to examine the optimized trace from \ours exploration. Figure~\ref{fig:trace} displays the initial four states and the corresponding three action-state transitions for the best reward path associated with the NCBI task~\citep{dougan2014ncbi} to extract disease entities. We highlight the domain insights by colors in both actions and states, where consistent coloring signifies analogous insights. Observably, from an initial human-composed prompt as $s_0$, \ours discovers various insightful error feedback (action) and effectively merges them into a refined prompt (state) with improved test performance. Over successive transitions, the definition of disease entities becomes increasingly refined, and biomedical-specific details are seamlessly integrated. The accumulation of this iterative process is reflected in the last state, $s_3$, which, infused with aggregated insights from its preceding path, manifests as an expert-level prompt, leading to a superior performance. 

We further annotate various quality aspects of optimized expert prompts, highlighting important perspectives on how expert prompts advance prompt engineering and provoke advanced task understanding of LLMs. As shown in Table~\ref{tab:ncbi_highlights} for the NCBI task and Appendix~\ref{sec:prompt_annotation} for all other tasks, in comparison with general human prompts and APE-optimized prompts, \ours prompts are typically more elaborate, offering comprehensive task instruction, which covers various diverse aspects, such as clarifying terminologies, guiding solutions, handling exceptional cases, etc. It is imperative to mention that while future research might explore prompt compression techniques~\citep{jiang2023llmlingua, yin2023did} to condense the expert prompt without sacrificing performance, the increased complexity of expert-level prompting naturally aligns with the advancement of contemporary state-of-the-art LLMs, enabling more sophisticated understanding of tasks and human requests.

\vspace{-5pt}
\section{Conclusion}
\vspace{-5pt}

In this paper, we introduce \ours, a novel prompt optimization framework capable of autonomously crafting expert-level prompts for a given task. Expert-level prompting distinguishes itself from traditional prompt engineering by its effectiveness of seamlessly integrating domain insights and closing the knowledge gap for domain experts. To achieve this, central to \ours is the novel perspective of viewing prompt optimization as a strategic planning problem, leveraging the power of MCTS planning to strategically and efficiently traverse the complex prompt space. \ours incorporates domain-specific knowledge from tasks into the newly generated prompts through a trial-and-error manner based on the self-reflection abilities of LLMs. We tested the \ours on 12 diverse tasks spanning three distinct domains. The prompts optimized by \ours consistently exhibited expert-level characteristics, enriched with domain-specific details and guidance. These prompts significantly outperformed both human-written, Chain-of-Thought prompting and other optimized method baselines. Further in-depth analyses revealed superior transferability, exploration efficiency, and quality for our expert prompts, paving the way for future prompt engineering to unlock the sophisticated task understanding of state-of-the-art LLMs.

\bibliography{main}
\bibliographystyle{iclr2024_conference}

\appendix

\clearpage
\newpage

\algnewcommand{\Inputs}[1]{%
  \State \textbf{Inputs:}
  \Statex \hspace*{\algorithmicindent}\parbox[t]{0.98\linewidth}{\raggedright #1}
}
\algnewcommand{\Initialize}[1]{%
  \State \textbf{Initialize:}
  \Statex \hspace*{\algorithmicindent}\parbox[t]{0.98\linewidth}{\raggedright #1}
}

\begin{algorithm}[h]
\centering
\caption{$\operatorname{\ours-MCTS}(s_0, p_\theta, r_\theta, p_\phi, d, L, \tau, c)$ 
\label{alg:mcts}}
\begin{minipage}{1\linewidth} 
\small
\begin{algorithmic}
    \Inputs{
    Initial prompt (state) $s_0$, state transition function $p_\theta$, reward function $r_\theta$, action generation function $p_\phi$, number of generated actions $d$, depth limit $L$, iteration number $\tau$, exploration weight $c$ (Equation~\ref{eq:uct})
    }
    \Initialize{
    State to action mapping $A : \mathcal S \mapsto \mathcal A$, children mapping $\text{ch} : \mathcal S \times \mathcal A \mapsto \mathcal S$, rewards $r : \mathcal S \times \mathcal A \mapsto \mathbb R$, \\
    State-action value function $Q : \mathcal S \times \mathcal A \mapsto \mathbb R$, visit-time counter $\mathcal{N} : \mathcal S \mapsto \mathbb N$
    }
    \For {$n \gets 0, \dots, \tau - 1$}
        \For {$t \gets 0, \dots, L - 1$}
            \If{$A(s_t)$ is not empty} \Comment{selection}
            \State $a_t \gets \arg\max_{a \in {A}(s_t)} \left( Q(s_t, a) + c\cdot \sqrt{\frac{\ln \mathcal{N}(s_t)}{\mathcal{N}(\text{ch}(s_t, a))}} \right)$ 
            \State $s_{t + 1} \gets \text{ch}(s_t, a_t)$, $r_t \gets r(s_t, a_t)$, $\mathcal{N}(s_{t}) \gets \mathcal{N}(s_{t}) + 1$ 
            \Else \Comment{expansion and simulation}
                \For {$i \gets 1, \dots, d$}
                \State Sample $a_t^i \sim p_\phi(a | s_t)$, $s_{t+1}^i \sim p_\theta(s | s_t, a_t^i)$, and $r_t^i \gets r_\theta(s_t, a_t^i)$
                \State Update $A(s_t) \gets \{a_t^i\}_{i=1}^d$, $\text{ch}(s_t, a_t^i) \gets s_{t+1}^i$, and $r(s_t, a_t^i) \gets r_t^i$
                \EndFor
            \State $a_t \gets \arg\max_{a^i_t \in {A}(s_t)} r_t^i (s_t, a^i_t) $ 
            \State $s_{t + 1} \gets \text{ch}(s_t, a_t)$, $r_t \gets r(s_t, a_t)$, $\mathcal{N}(s_{t}) \gets \mathcal{N}(s_{t}) + 1$ 
            \EndIf
            \State \algorithmicif\ {$s_{t+1}$ is an early-stopping state}\ \algorithmicthen\ \textbf{break}

        \EndFor
        \State $T \gets$ the actual number of steps
        \For {$t \gets T - 1, \dots, 0$} \Comment{back-propagation}
            \State Update $Q(s_t, a_t)$ with $\{r_t, r_{t+1}, \dots, r_L\}$ based on Equation~\ref{eq:backpropagate}
        \EndFor
    \EndFor
\end{algorithmic}
\end{minipage}
\end{algorithm}

\clearpage
\section{More Experiment Details}
\label{sec: experimental setting}

\subsection{Input Formulation}
\label{input_and_dataset}

The normal model input is composed of the following components:

\label{input_format}
\begin{center}
    Prompt + \textcolor{gray}{Task Prefix} + Question + \textcolor{gray}{Task Suffix} + \textcolor{gray}{Answer Format}
\end{center} 

``Prompt'' is the optimization target. ``Task Prefix'' (Optional) is the task-specific background intro (For example, a table of background data in the Penguins). ``Question'' is the main body of the task's question. ``Task Suffix'' (Optional) includes the options (For example, yes/no, entailment/non-entailment, or A, B, C, D in tasks with multiple choices). ``Answer Format'' (Optional) is designed for answer caption from the model's response. Examples of the task input are in Appendix~\ref{input_examples}.

The meta formats and prompts, as explained in Section~\ref{sec:framework}, are in Appendix~\ref{meta_formats}.\\

\begin{wraptable} {r}{0.48\linewidth}
\vspace{-10pt}
\caption{ {Data split}}
\vspace{-5pt}
\label{tab: split} 
\definecolor{Gray}{gray}{0.90}
\newcolumntype{a}{>{\columncolor{Gray}}c}
\centering
\resizebox{0.8\linewidth}{!}{%
\begin{tabular}{@{}lp{1cm}c@{}}
\toprule
Task    & Train & Test \\ \midrule
\textbf{Bigbench}       &       &   \\
Penguins        &  70 &  79    \\
Geometry &    150   & 200\\
Epistemic & 500 & 500 \\
Object counting & 300 & 500\\
Temporal& 300 & 500 \\
causal judgement & 90 &100 \\
\midrule
\textbf{Domain Knowledge} && \\
NCBI& 2000 & 940 \\
Biosses & 60 & 40 \\
Med QA & 2000& 500 \\
\midrule
\textbf{General NLP} &&  \\
Subj &400 & 1000 \\
TREC & 400 & 500 \\
CB & 125 & 56 
\\ \bottomrule
\end{tabular}
}
\vspace{-10pt}
\end{wraptable}

\subsection{Data Split}

For datasets with predefined testing sets, we directly use them as our testing set. When these exceed 1,000 examples, we sample 1000 from them. If no default testing set is provided, we shuffle the data and allocate approximately half for testing purposes. We then sample a subset from the remaining data as the training set. From this training set, a held-out subset is sampled for reward calculation with a default size of 150. If the training set is smaller than 150 or very large, the subset will range between 60 to 200 examples accordingly. The data split details are in Table~\ref{tab: split}.

\subsection{More Implementation Details}
\label{prompt_agent_implement}

\noindent \textbf{PromptAgent (Ours).} 
\ours performs MCTS planning within the prompt space, requiring both terminal state conditions and a reward function. A terminal state is achieved when the path length hits \textit{depth\_limit}. The reward function is determined by the base model's performance on the held-out set. For computational efficiency to avoid unnecessary exploration, we also apply an early-stopping method after depth is larger than 2: if the state's reward is less than a \textit{min\_threshold} or larger than a \textit{max\_threshold}, we then reach an early-stopping state. Specifically, \textit{min\_threshold} is the average of the rewards of its parent node and the root node, while \textit{max\_threshold} is the maximum of all the current nodes, which encourages shorter paths. We now further illustrate the details of Algorithm~\ref{alg:mcts}. 

\begin{enumerate}
    \item \textbf{Initialization}. The \ours-MCTS algorithm starts with an initial prompt as the root node. For BBH tasks, we directly adopt the task ``description'' from the original datasets as the initial prompts, except that \textit{Object Counting}'s default description doesn't follow the format of instruction. We crafted the initial prompts for the rest of the tasks according to their task objectives or question-answer formats. The root node will be evaluated to obtain the reward before the first expansion. 
    \item \textbf{MCTS Iterations}. The agent will perform 12 MCTS iterations. During the selection step, starting from the root node, the best child node will be added to the path according to its UCT value (Equation~\ref{eq:uct}), and the exploration weight $c$ in UCT is 2.5. During the expansion step, \textit{expand\_width} batches (\textit{batch\_size} is $5$) of examples will be sampled from the training set, and each batch will be fed to the base model to collect the errors. If there is no error, this sample-forward loop will iterate until an error is found. The errors will be formatted using \textbf{\texttt{error\_string}} (illustrated in Table~\ref{error_string}) and inserted into \textbf{\texttt{error\_feedback}} (illustrated in Table~\ref{error_feedback}, Meta-prompt 1 in Figure~\ref{fig:framework}) to summarize errors by the optimizer. \textbf{\texttt{state\_transit}} prompt (illustrated in Table\ref{state_transit}, Meta-prompt 2 in Figure~\ref{fig:framework}) contains the expanding node's prompt, the trajectory of prompts (list of prompts from the root of the expanding node on the currently selected path), and the error summarization, which is fed into the optimizer to generate \textit{num\_samples} new prompts (nodes). The new nodes will be evaluated and added as the expanding node's children if they are not terminal nodes. Each expansion will generate \textit{expand\_width} $\times$ \textit{num\_samples} new prompts. The simulation step will recursively expand the last node in the path and pick the one with the highest reward to add to the path. When the last node satisfies the terminal condition or early-stopping condition, the simulation is stopped. During the back-propagation, from the last node to the root, the cumulative rewards (the sum of rewards from the node to the leaf/terminal node) will be appended to the node's cumulative reward list, the average of which will be the node's $Q$ (Equation~\ref{eq:backpropagate}). We have three hyperparameter settings: Standard, Wide, and Lite in Table~\ref{tab:hyperparameter}. In the Standard and Lite experiments, both have an \textit{expand\_width} of 3 and \textit{num\_samples} of 1, but their \textit{depth\_limit} are 8 for Standard and 4 for Lite. Wide experiment has \textit{expand\_width} is $3$ and $num\_samples=2$ to generate more nodes in each expansion step, but with a \textit{depth\_limit} of 6 to limit the total number of explored prompts. We select the best setting for each task based on the final rewards.
    \label{output_strategt}
    \item \textbf{Output strategy}. Each MCTS iteration will output one path from the root node to the leaf node, and there are tens of nodes generated after the searching process. We select the path with the highest average reward, then pick the prompt with the highest reward in the path as the final output prompt. We employ this strategy because the path with the highest average reward represents the best overall search trajectory, and also, the best prompt might not always be the last node on the optimal path, given that it may be a terminal state by reaching the depth limit.
\end{enumerate}
\label{sec: hyperparameter}
\begin{table}[h]
\caption{Hyperparameter settings for PromptAgent Experiments}
\label{tab:hyperparameter} 
\definecolor{Gray}{gray}{0.90}
\newcolumntype{a}{>{\columncolor{Gray}}c}
\centering
\resizebox{0.5\linewidth}{!}{%
\begin{tabular}{@{}lccc@{}}
\toprule
        Experiment Name & Standard & Wide & Lite\\ \midrule
        \textit{depth\_limit}  & 8 & 6 & 4\\
        \textit{expand\_width}  & 3 & 3 & 3\\
        \textit{num\_samples}  & 1 & 2 & 1\\\bottomrule
\end{tabular}
}
\end{table}

\subsection{Baselines Implementation Details}
\label{search_baselines}

We illustrate the details for various baselines in our experiments.

\noindent \textbf{Monte Carlo (MC).}
MC performs one-step sampling multiple times and selects the best one as the optimized prompt. It uses the same prompt sampling method as \ours, but limits the searching depth to one. In the search ablation study, we sampled 72 new prompts in each task.

\noindent \textbf{Beam Search (Beam).}
Beam also uses the same expand function as \ours. Each node, except the root, will be expanded into 3 new nodes, and the beam width is 3, meaning that there will be 9 nodes in each depth of the search tree, and the best 3 nodes will be kept for the next expansion. The root will be expanded into 9 new nodes. The search depth is 8, so there will be 72 nodes or new prompts in total. 

\noindent \textbf{Greedy Search (Greedy).}
Greedy is based on the Beam Search, but the beam width is one, so the algorithm turns into a depth-first greedy search. We conducted 2 experiments, Greedy-S and Greedy-L, in Figure~\ref{fig:efficiency}, with the same search depth of 8 but different expand widths. The Greedy-S's expand width is 3, and it has 34 prompts in total. The Greedy-L has an \textit{expand\_width} of 9 and 72 nodes in total, which is also referred to as the Greedy baseline in Table~\ref{tab:ablation}.

\noindent \textbf{APE~\citep{zhou2022large}.}  
We employ the iterative APE with one iteration as our baseline, as suggested by the original paper~\citep{zhou2022large}. When generating new prompts, a mini-batch comprising 5 data pieces is sampled as Input-Output examples for APE. 
Specifically, for \textbf{Initial Proposal Step}, by default, 10 data batches are sampled, with each batch being used to generate 10 new prompts. This results in a total of 100 candidate prompts during the initial step. (Due to the longer processing time of Med QA, only 25 candidates are generated for it in this phase.) Subsequently, the five prompts with the highest evaluation scores are chosen for the iterative proposal step. For \textbf{Iterative Proposal Step}, similar to the initial phase, 10 batches of data are sampled for each proposed prompt, resulting in a total of 50 candidate prompts in this step. Following this, the prompt with the top evaluation score is chosen as the optimized prompt.

\clearpage

\subsection{Meta Formats}
\label{meta_formats}
In this section, we present the full formats for meta-prompts used in the \ours. ``input\_format'' is the actual input of the base model given a question. ``error\_string'' represents the format of each error example. ``error\_feedback'' includes several error examples and guides the optimizer model to collect the error feedback. ``state\_transit'' guides the optimizer model to make state transitions (generate new prompts), which includes the information of error examples and the sequence of prompts in the selected path, which is the ``trajectory\_prompts''.\\

\begin{table}[h]
\caption{Meta Formats.}
\label{tab: meta_prompt} 
\definecolor{Gray}{gray}{0.90}
\newcolumntype{a}{>{\columncolor{Gray}}c}
\centering
\resizebox{0.9\linewidth}{!}{%
\begin{tabular}{@{}lp{10cm}}
\toprule
Format Name    & Meta Format  \\ 
\midrule
input\_format    & \{prompt\}\par\{task\_prefix\}\par\{question\}\par\{task\_suffix\}\par\{answer\_format\}  \\ 
\midrule
\label{error_string}
error\_string    & \textless\{index\}\textgreater \par The model's input is:\par \{question\}\par  \vspace{3pt} The model's response is: \par \{response\}\par \vspace{3pt} The correct label is: \{label\}\par  The model's prediction is \{prediction\}  \\ 
\midrule
\label{error_feedback}
error\_feedback       &    I'm writing prompts for a language model designed for a task. \par  \vspace{3pt} My current prompt is:\par \{cur\_prompt\}\par \vspace{3pt} But this prompt gets the following examples wrong:\par \{error\_string\}\par  \vspace{3pt} For each wrong example, carefully examine each question and wrong answer step by step, provide comprehensive and different reasons why the prompt leads to the wrong answer. At last, based on all these reasons, summarize and list all the aspects that can improve the prompt.          \\
\midrule
\label{state_transit}
state\_transit         &     I'm writing prompts for a language model designed for a task.\par  \vspace{3pt} \par My current prompt is:\par \par \{cur\_prompt\}\par \vspace{3pt} But this prompt gets the following examples wrong:\par\par \{error\_string\}\par \vspace{3pt} Based on these errors, the problems with this prompt and the reasons are:\par\par \{error\_feedback\}\par \vspace{3pt} There is a list of former prompts including the current prompt, and each prompt is modified from its former prompts:\par\par \{trajectory\_prompts\}\par \vspace{3pt} Based on the above information, please write \{steps\_per\_gradient\} new prompts following these guidelines:\par\par 1. The new prompts should solve the current prompt's problems.\par\par 2. The new prompts should consider the list of prompts and evolve based on the current prompt.\par\par 3. Each new prompt should be wrapped with \textless{}START\textgreater and \textless{}END\textgreater{}.\par  \vspace{3pt}  The new prompts are: \\  \bottomrule
\end{tabular}
}
\end{table}

\clearpage
\section{Task Input Examples}
\label{input_examples}

In this section, we show some input examples in several tasks for the base model. Specifically, our tasks fall into three categories: multi-choice selection, name entity recognition, and direct answer matching. As representative examples, we select \textit{Penguins in A Table}, \textit{NCBI}, and \textit{Subjective} to illustrate the input format.

\begin{figure}[h]
    \centering
    \includegraphics[width=0.95\linewidth]{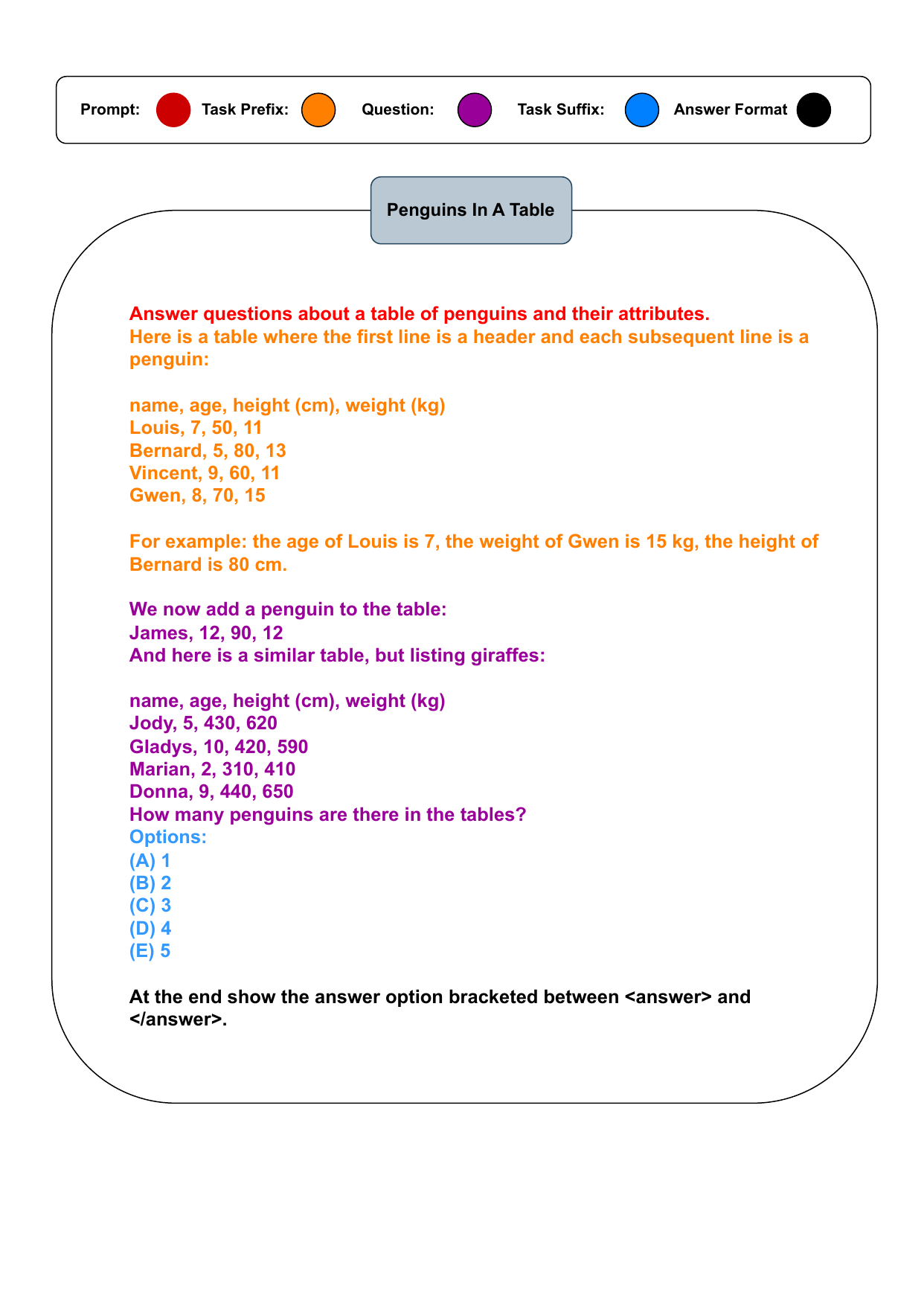}
    \caption{Input format of \textit{Penguins in A Table} task.}
    \label{fig:input_format_penguins}
\end{figure}

\begin{figure}
    \centering
    \includegraphics[width=1\linewidth]{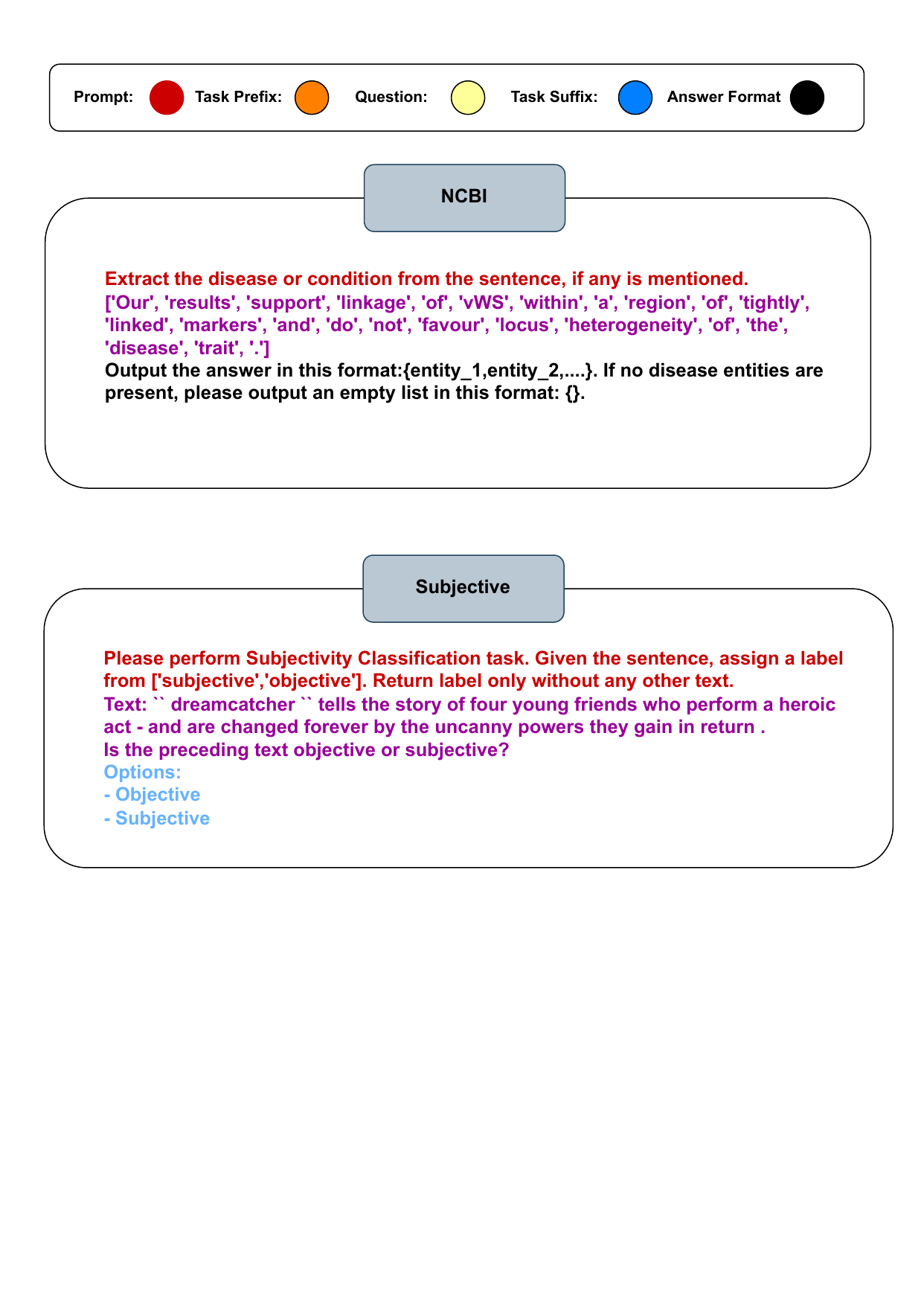}
    \caption{Input formats of \textit{NCBI} and \textit{Subjective} task.}
    \label{fig:enter-label}
\end{figure}

\clearpage
\section{Convergence observation details}
\label{sec: Convergence observation}
\begin{figure}[htbp]
    \centering
    \begin{subfigure}{0.4\textwidth}
        \includegraphics[width=\textwidth]{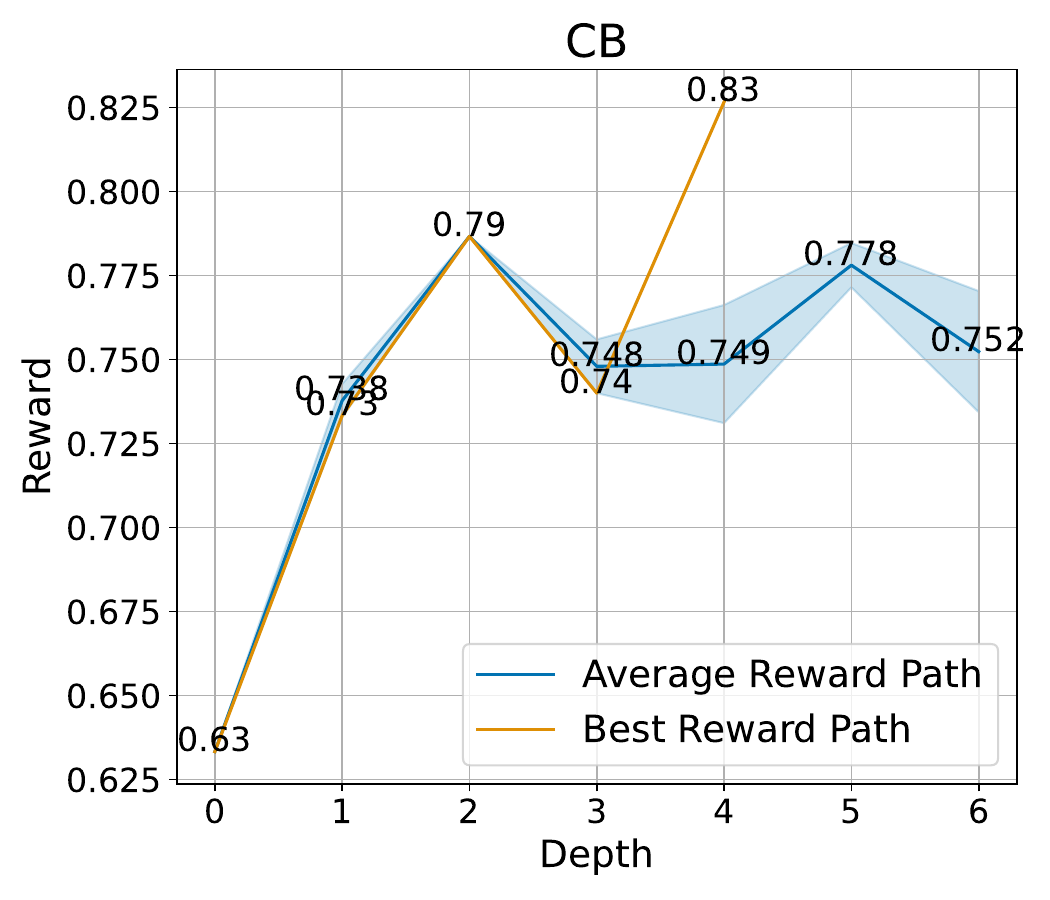}
    \end{subfigure}%
    \hspace{3em} 
    \begin{subfigure}{0.4\textwidth}
        \includegraphics[width=\textwidth]{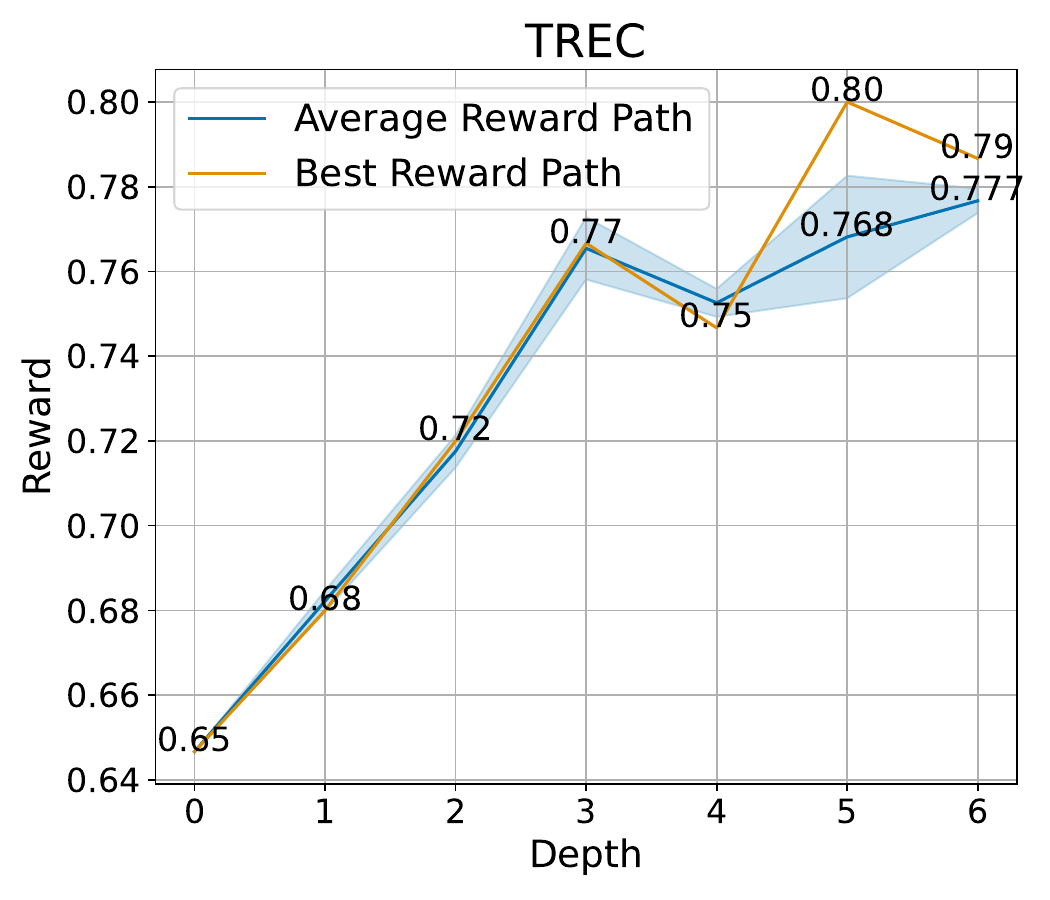}
    \end{subfigure}
    

    \begin{subfigure}{0.4\textwidth}
        \includegraphics[width=\textwidth]{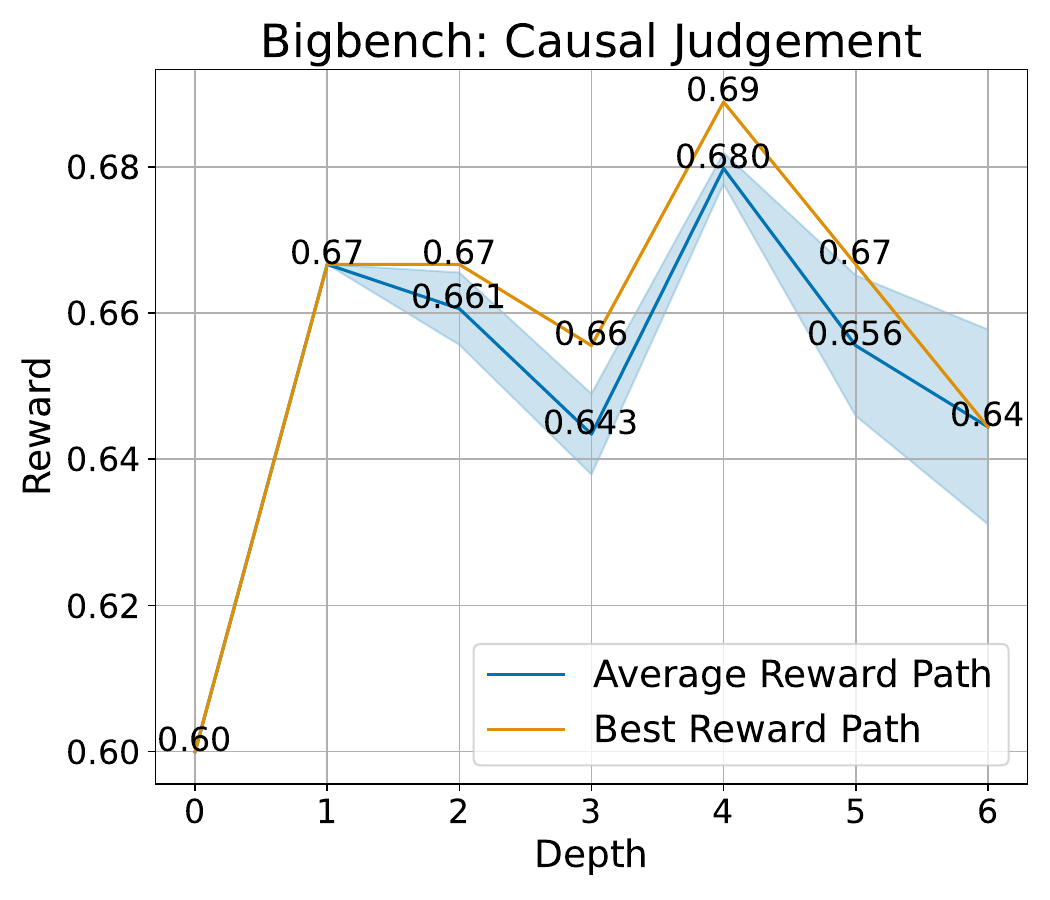}
    \end{subfigure}%
    \hspace{3em} 
    \begin{subfigure}{0.4\textwidth}
        \includegraphics[width=\textwidth]{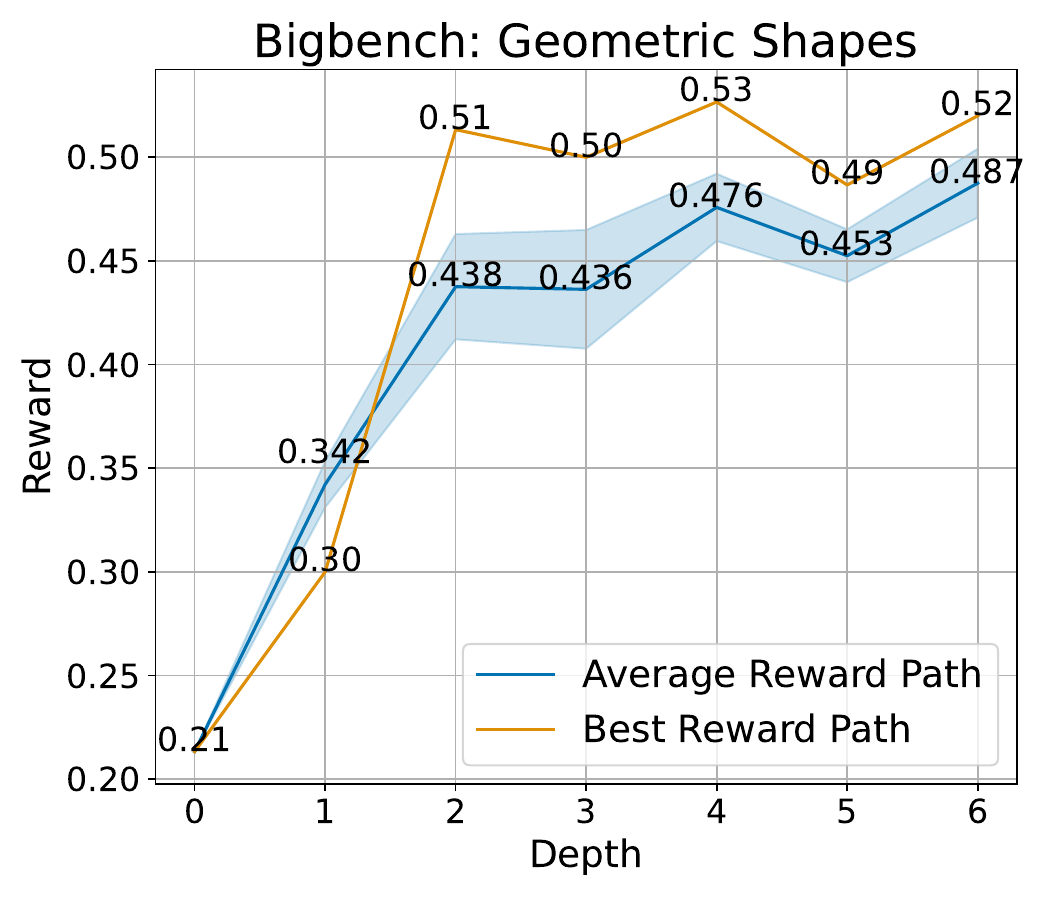}
    \end{subfigure}
    

    \begin{subfigure}{0.4\textwidth}
        \includegraphics[width=\textwidth]{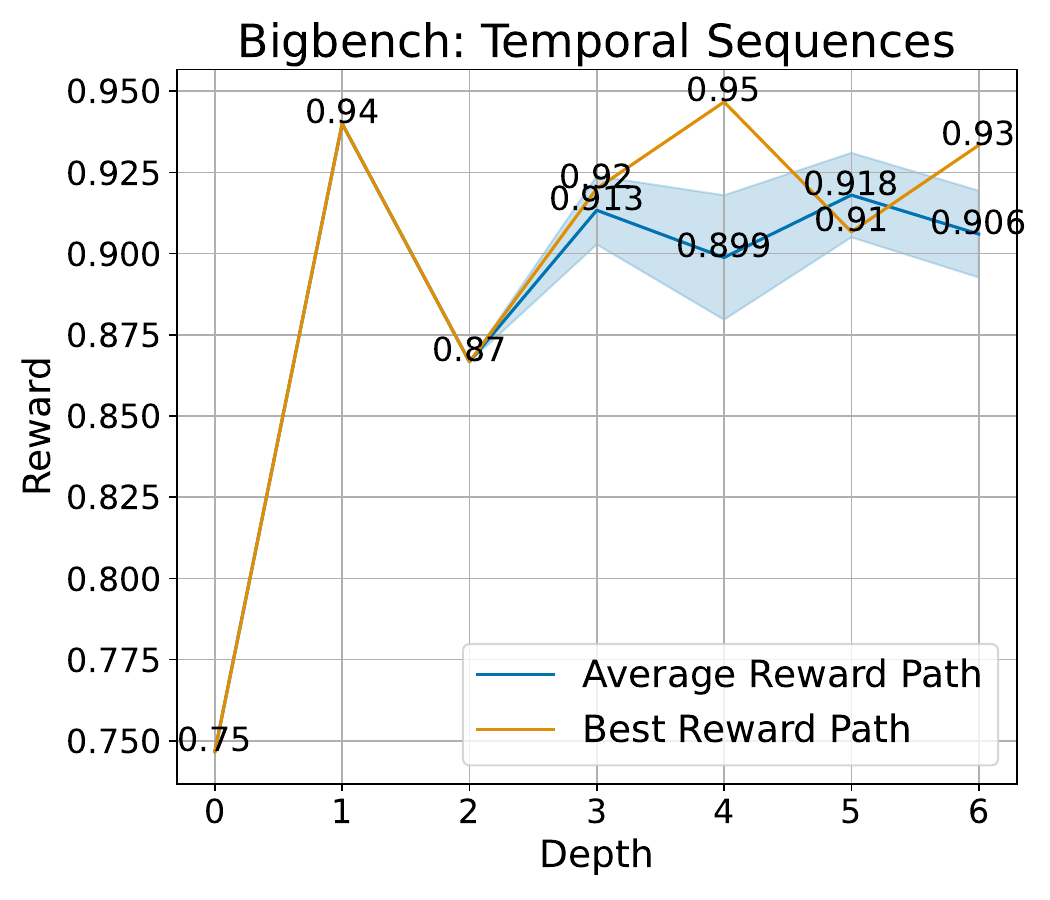}
    \end{subfigure}%
    \hspace{3em} 
    \begin{subfigure}{0.4\textwidth}
        \includegraphics[width=\textwidth]{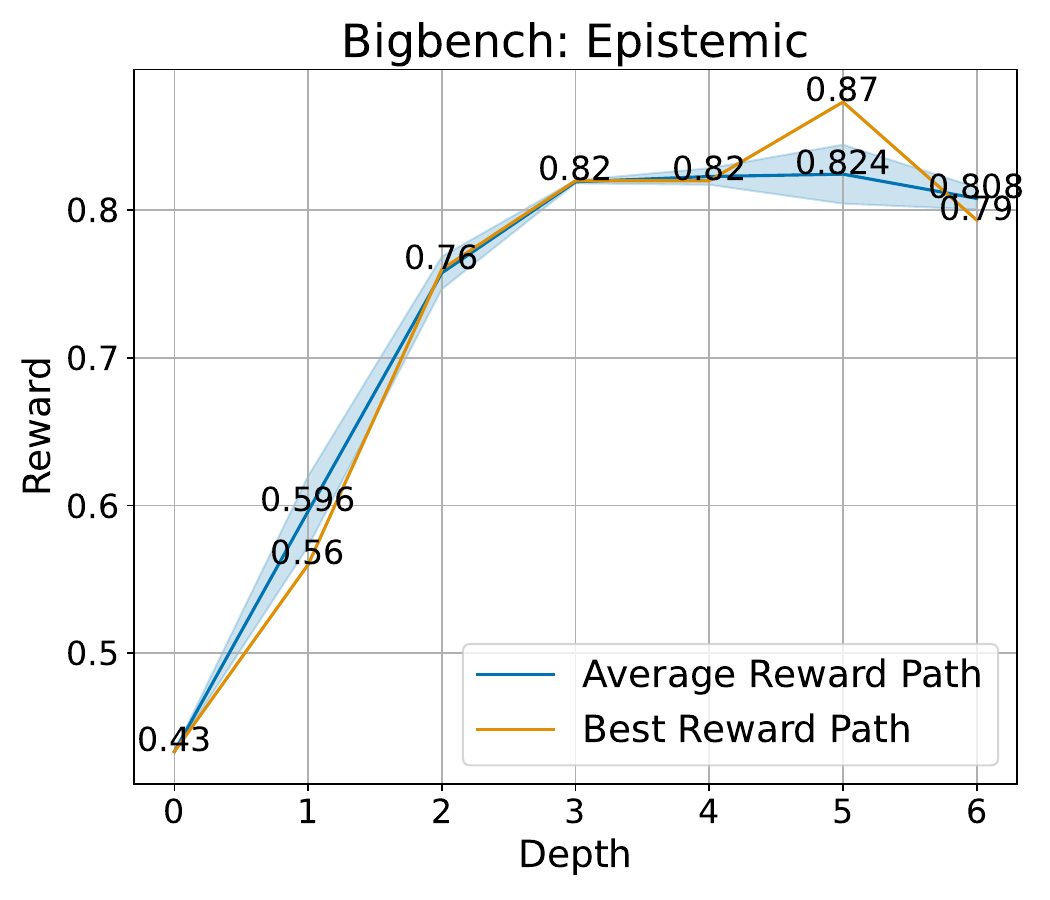}
    \end{subfigure}
    \vspace{0.2em} 

    \begin{subfigure}{0.4\textwidth}
        \includegraphics[width=\textwidth]{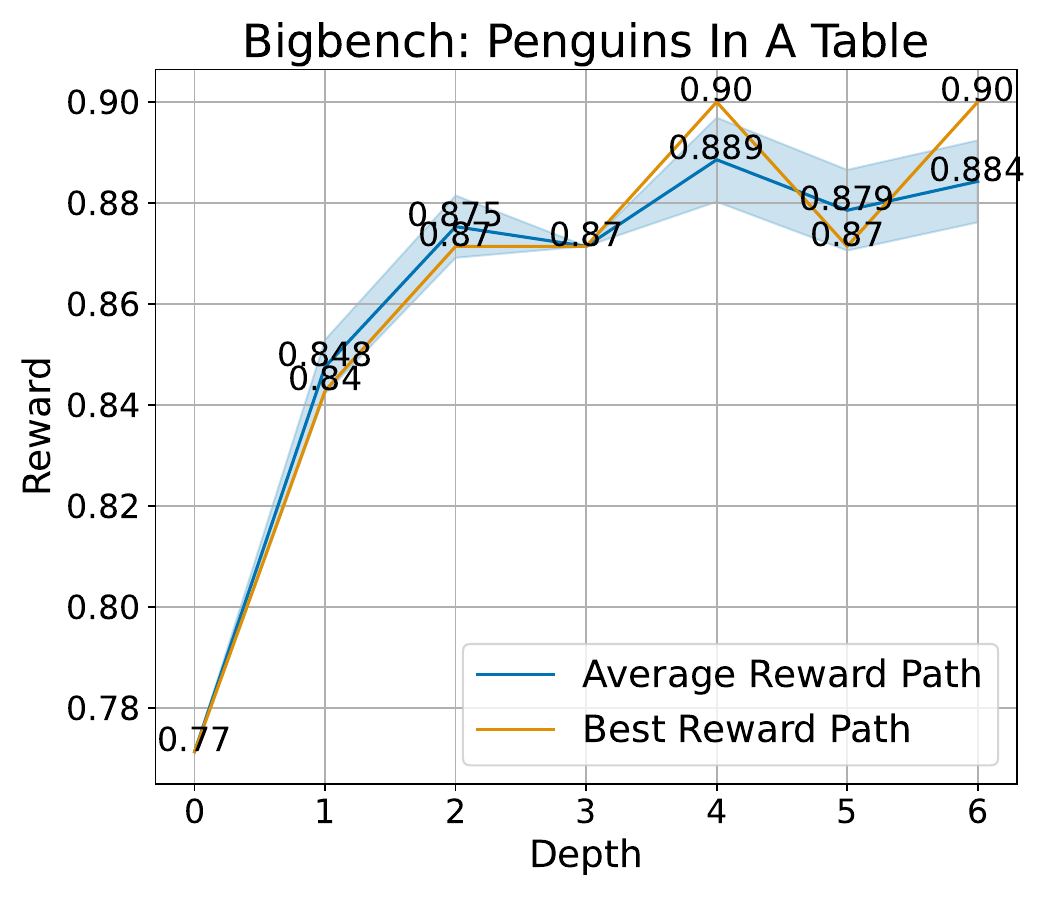}
    \end{subfigure}%
    \hspace{3em} 
    \begin{subfigure}{0.4\textwidth}
        \includegraphics[width=\textwidth]{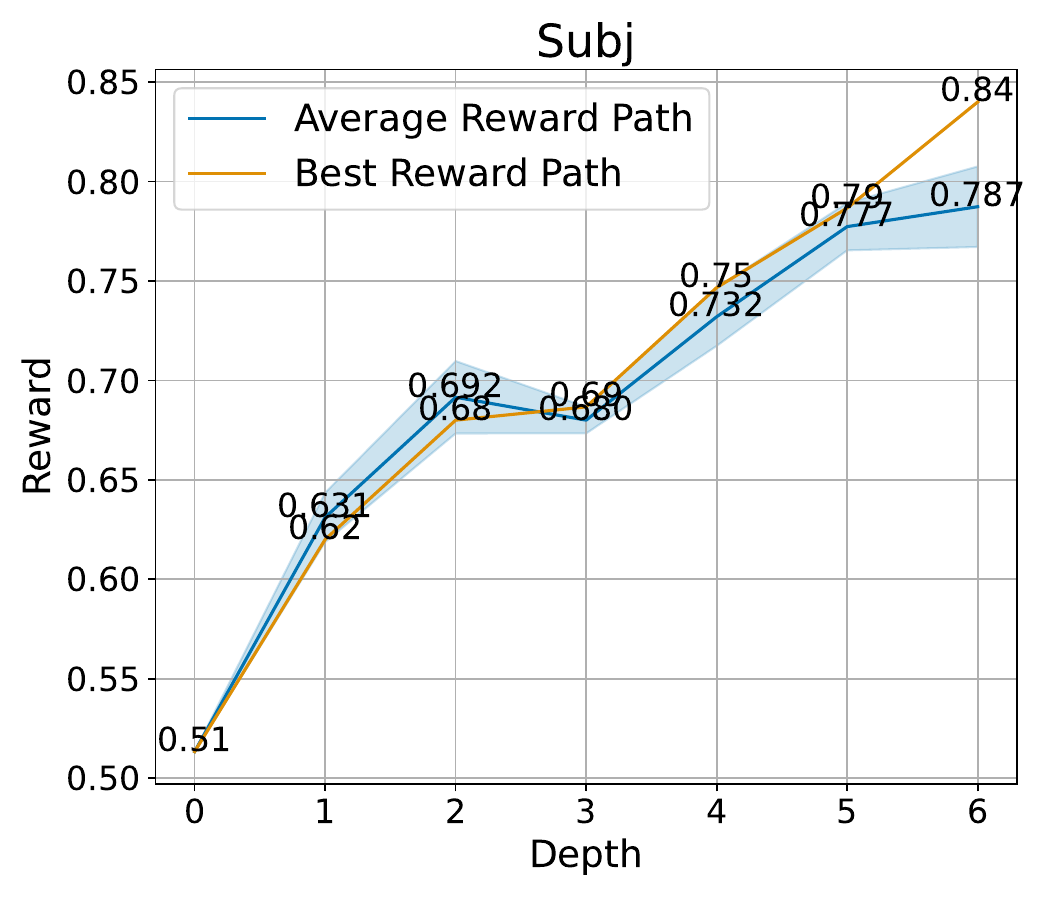}
    \end{subfigure}

    \caption{Convergence plots with the ``Wide'' setting. $\textit{expand\_width}=3$, $\textit{num\_samples}=2$, and $\textit{depth\_limit}=6$. The Average Reward Path is the average reward of paths, and the blue area is the variance. The Best Reward Path is the path with highest average reward,  where the best node is selected as the node with highest reward on the Best Reward Path.}
    \label{fig:mygraphs}
\end{figure}

\begin{figure}[htbp]
    \centering
    \begin{subfigure}{0.4\textwidth}
        \includegraphics[width=\textwidth]{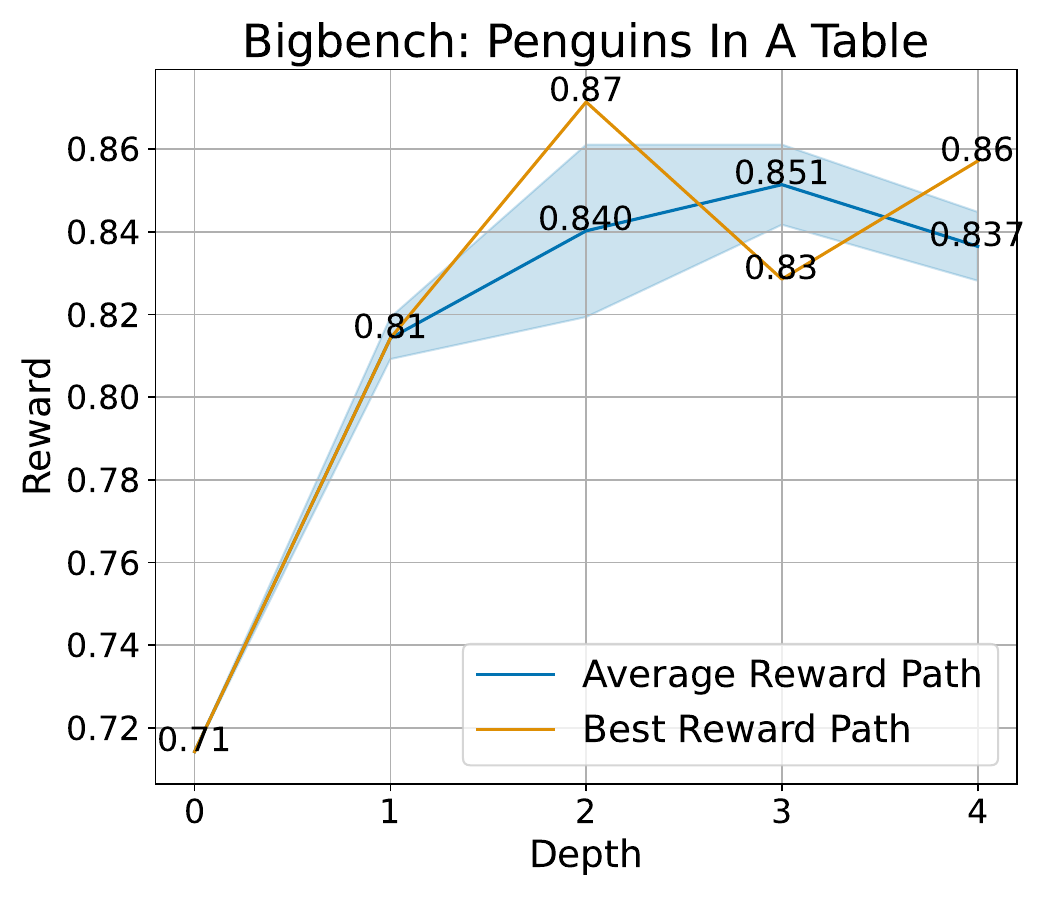}
    \end{subfigure}%
    \hspace{3em} 
    \begin{subfigure}{0.4\textwidth}
        \includegraphics[width=\textwidth]{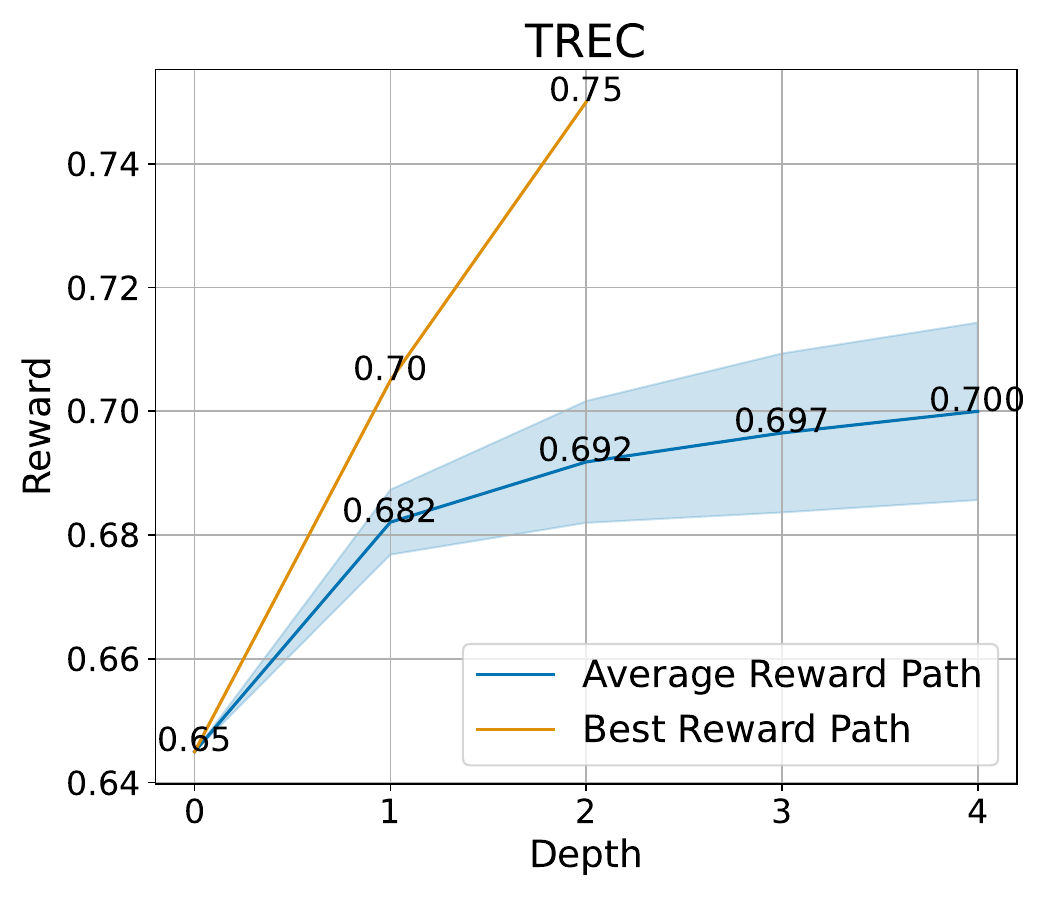}
    \end{subfigure}
    
    \vspace{0.6em} 

    \begin{subfigure}{0.4\textwidth}
        \includegraphics[width=\textwidth]{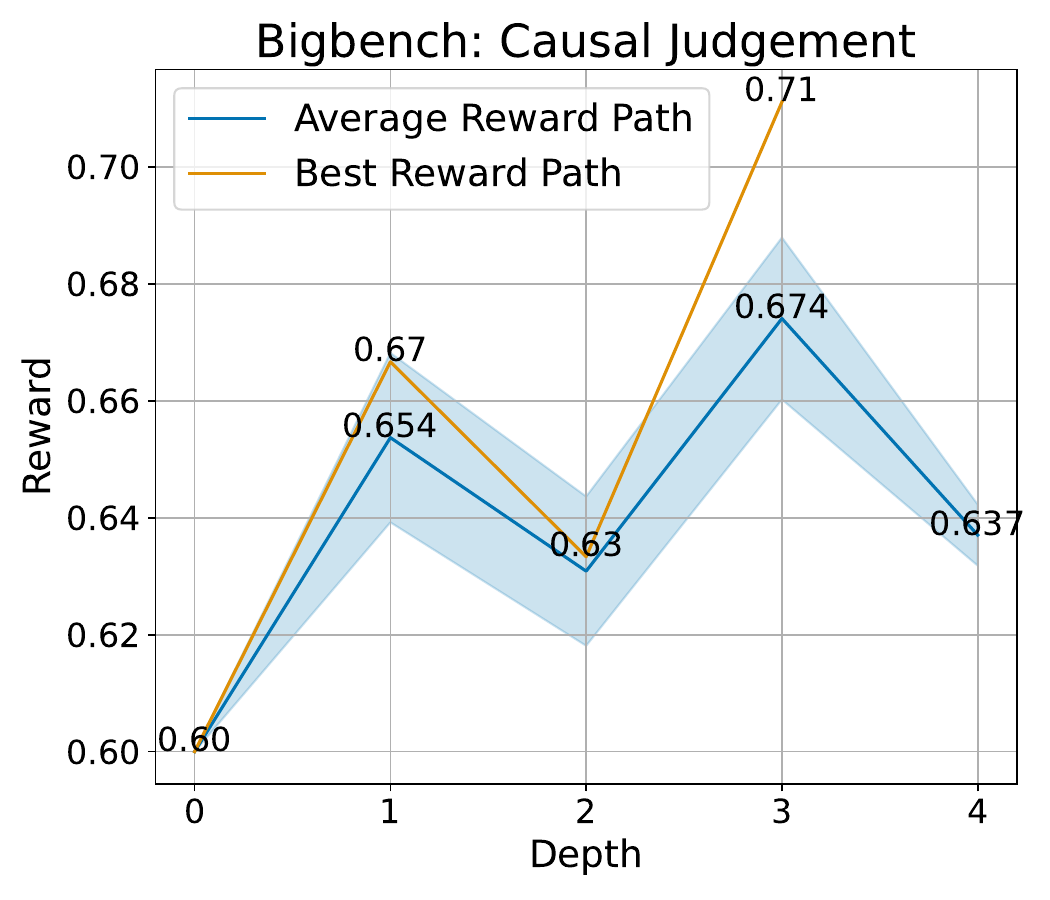}
    \end{subfigure}%
    \hspace{3em} 
    \begin{subfigure}{0.4\textwidth}
        \includegraphics[width=\textwidth]{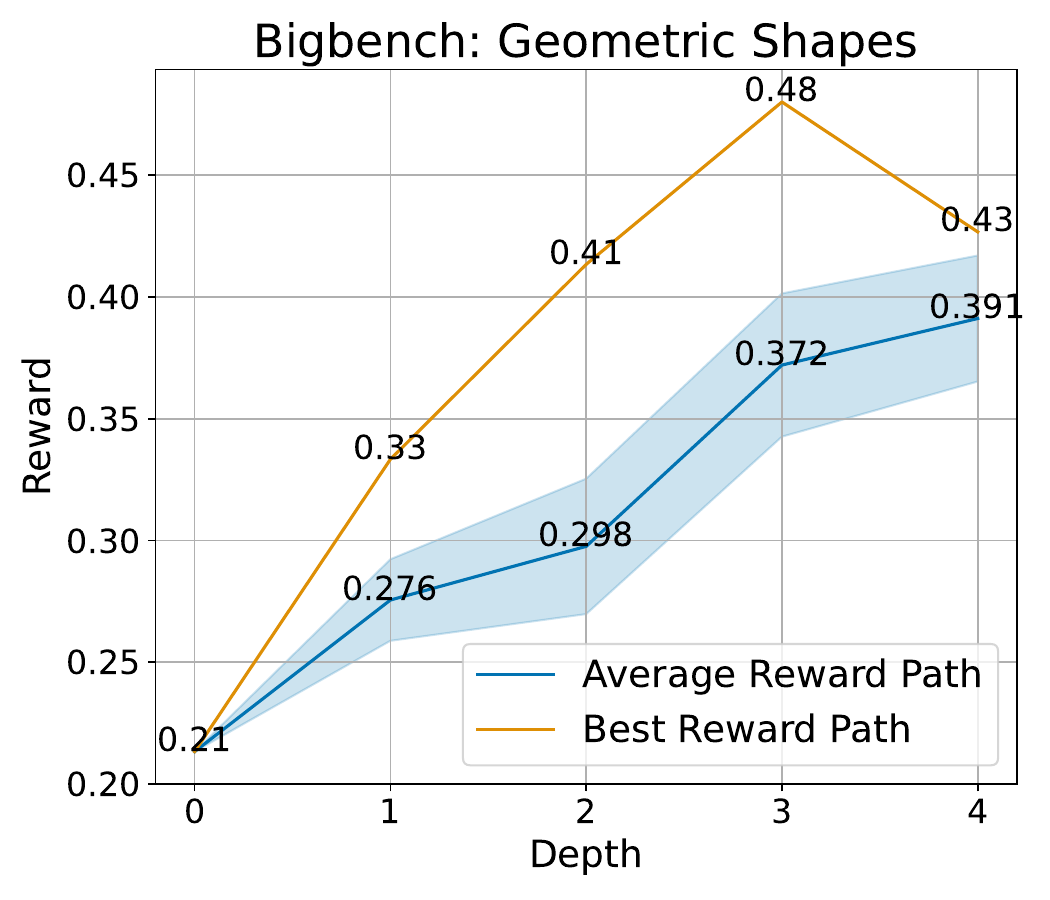}
    \end{subfigure}
    
    \vspace{0.6em} 

    \begin{subfigure}{0.4\textwidth}
        \includegraphics[width=\textwidth]{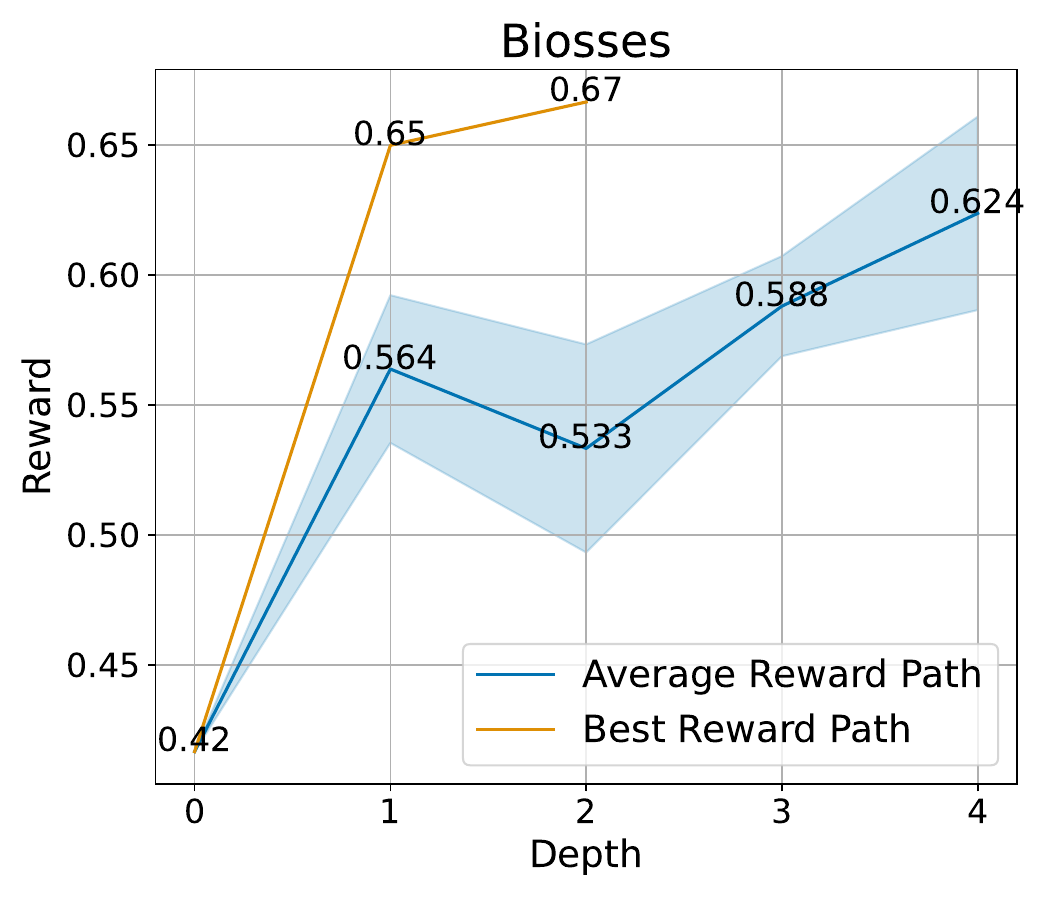}
    \end{subfigure}%
    \hspace{3em} 
    \begin{subfigure}{0.4\textwidth}
        \includegraphics[width=\textwidth]{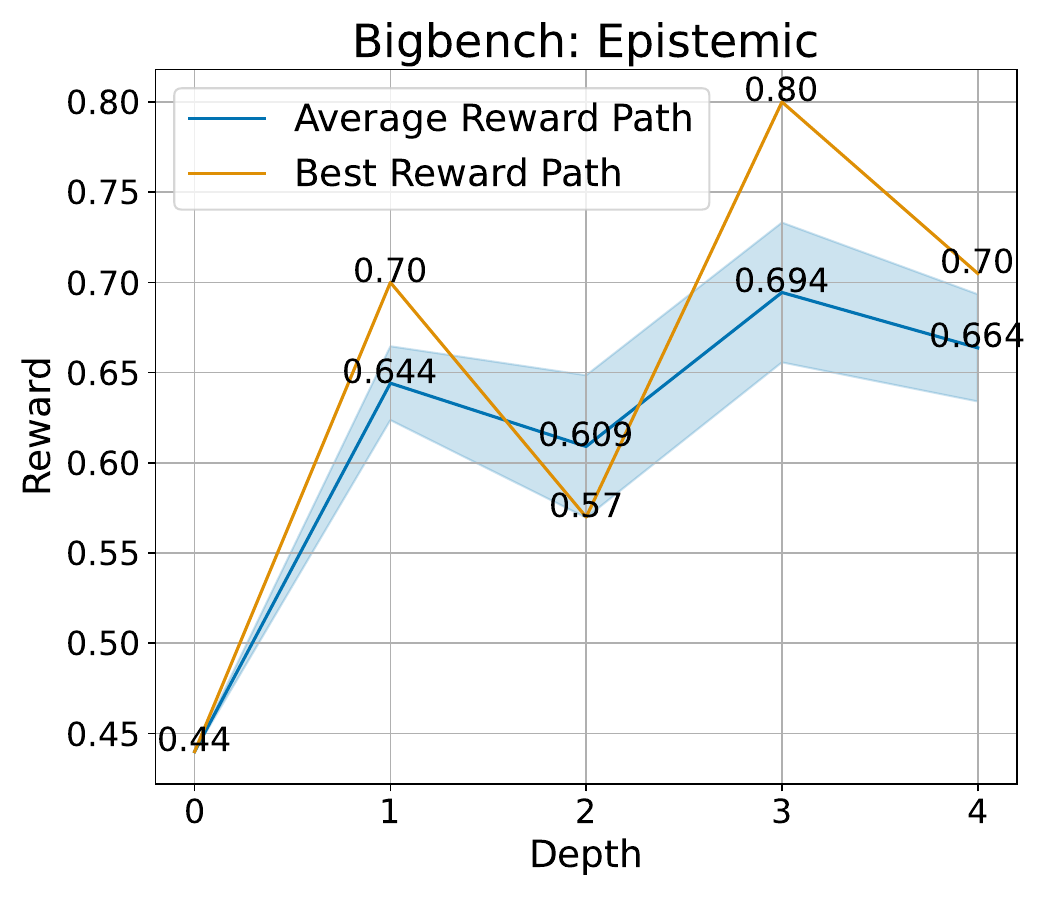}
    \end{subfigure}
    \vspace{0.6em} 

    \begin{subfigure}{0.4\textwidth}
        \includegraphics[width=\textwidth]{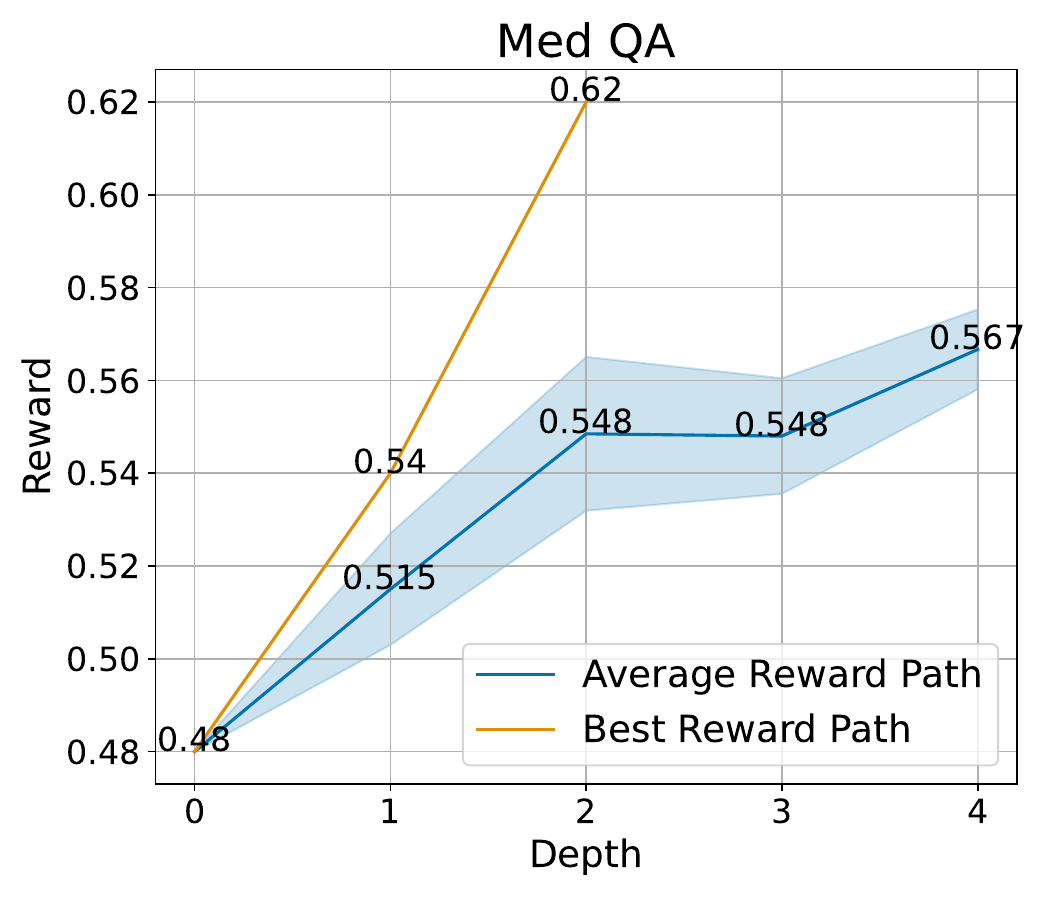}
    \end{subfigure}%
    \hspace{3em} 
    \begin{subfigure}{0.4\textwidth}
        \includegraphics[width=\textwidth]{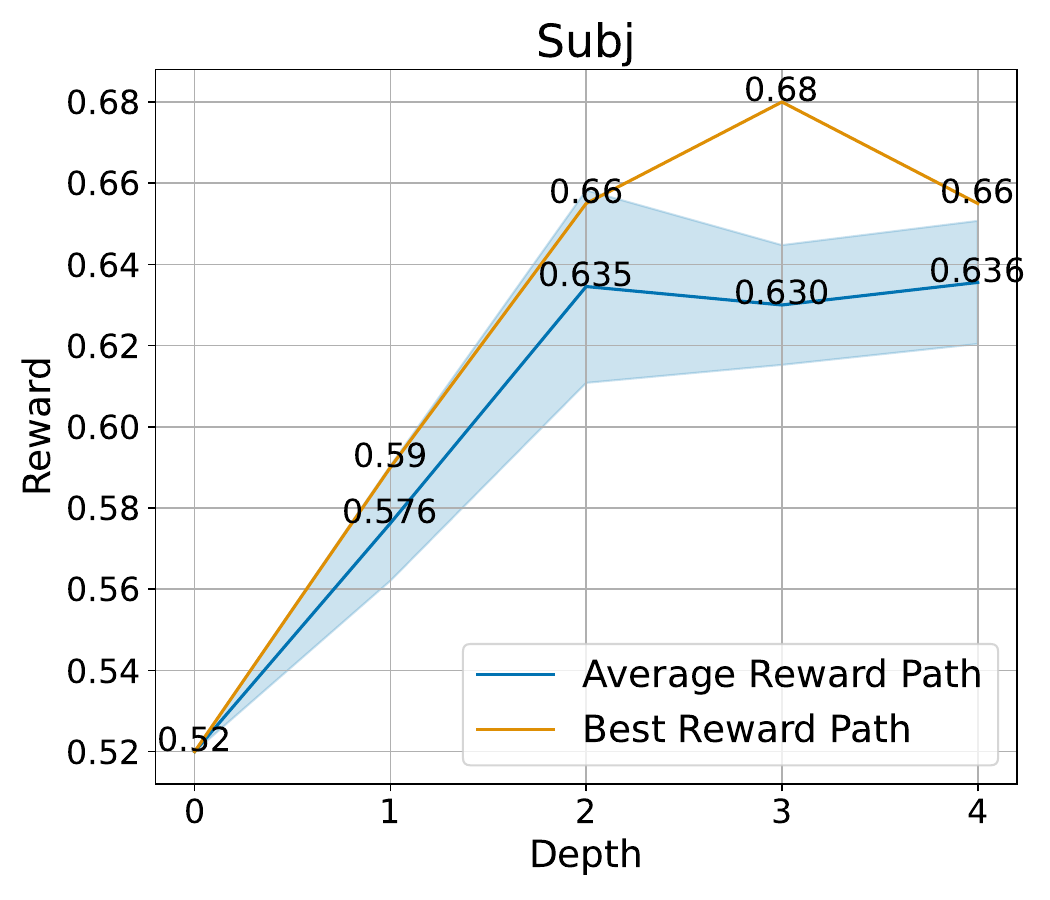}
    \end{subfigure}
    \vspace{0.6em} 

    \caption{Convergence plots with the ``Lite'' setting. $\textit{expand\_width}=3$, $\textit{num\_samples}=1$, and $\textit{depth\_limit}=4$. The Average Reward Path is the average reward of paths, and the blue area is the variance. The Best Reward Path is the path with highest average reward, where the best node is selected as the node with highest reward on the Best Reward Path.}
    \label{fig:mygraphs}
\end{figure}

\begin{figure}[htbp]
    \centering
    \begin{subfigure}{0.4\textwidth}
        \includegraphics[width=\textwidth]{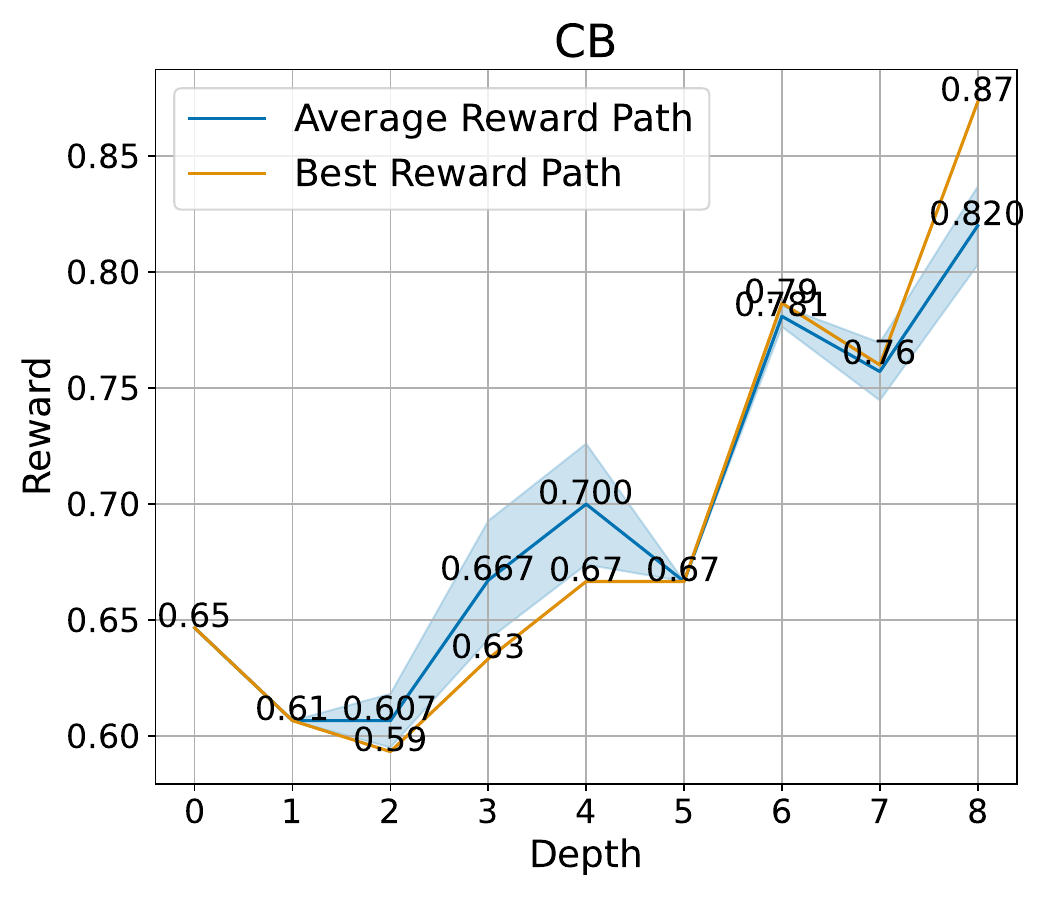}
    \end{subfigure}%
    \hspace{3em} 
    \begin{subfigure}{0.4\textwidth}
        \includegraphics[width=\textwidth]{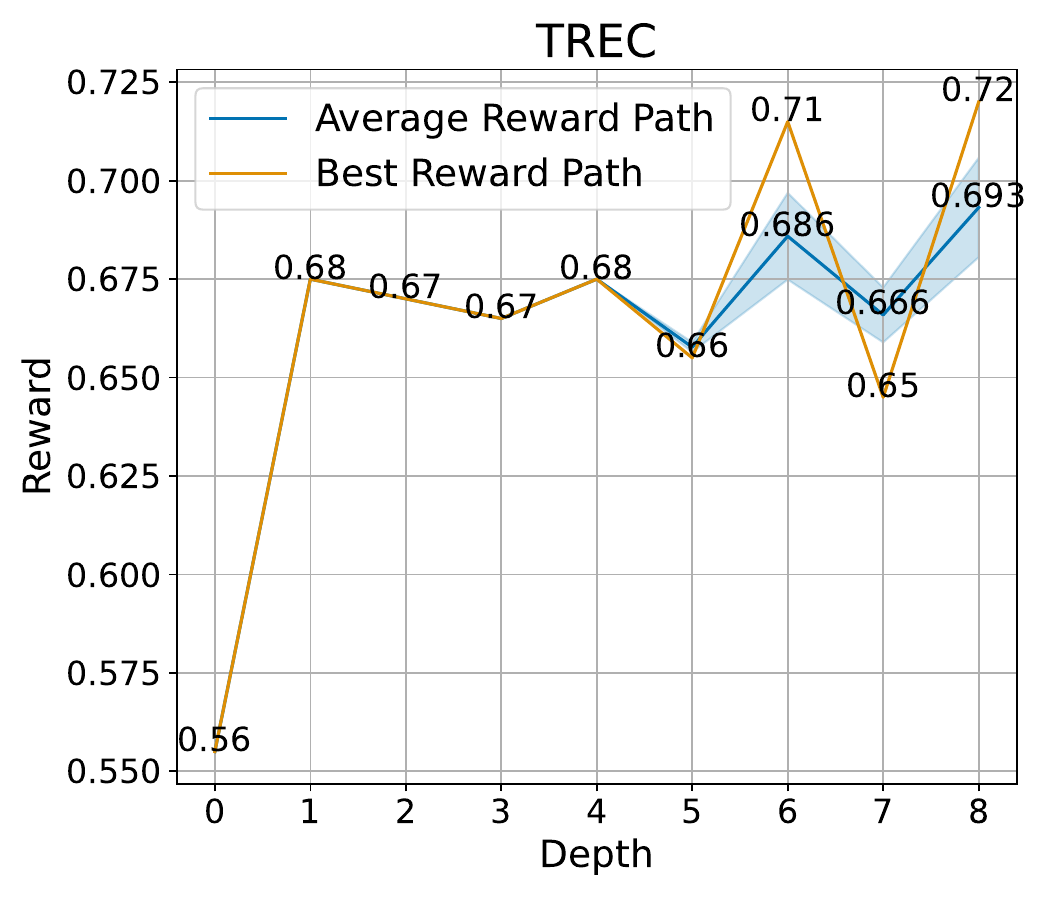}
    \end{subfigure}
    
    \vspace{0.6em} 

    \begin{subfigure}{0.4\textwidth}
        \includegraphics[width=\textwidth]{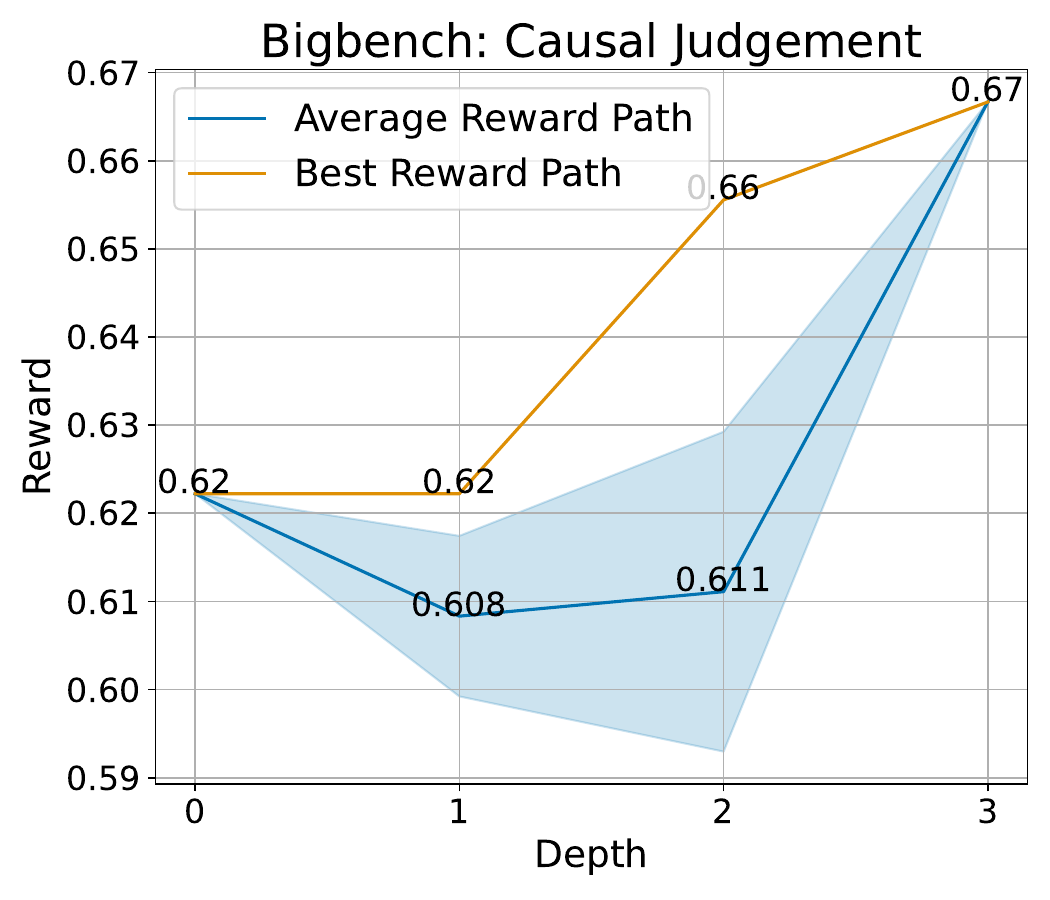}
    \end{subfigure}%
    \hspace{3em} 
    \begin{subfigure}{0.4\textwidth}
        \includegraphics[width=\textwidth]{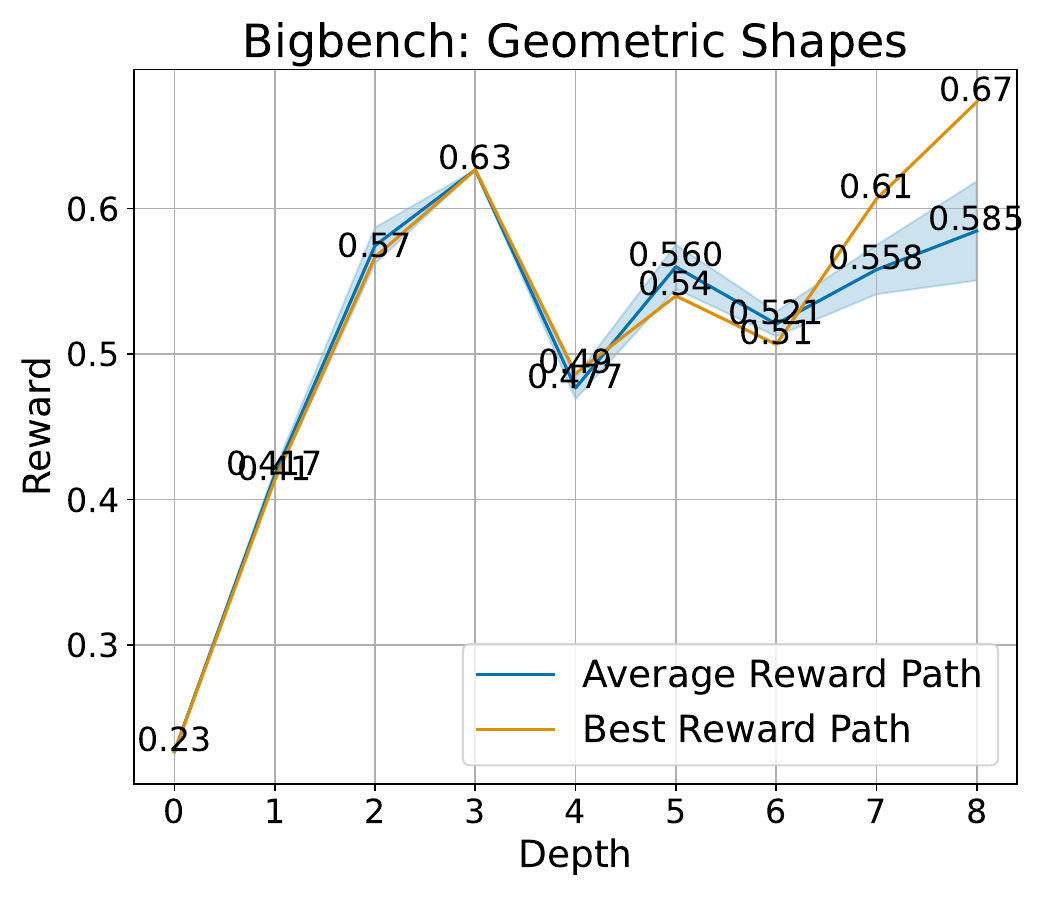}
    \end{subfigure}
    
    \vspace{0.6em} 

    \begin{subfigure}{0.4\textwidth}
        \includegraphics[width=\textwidth]{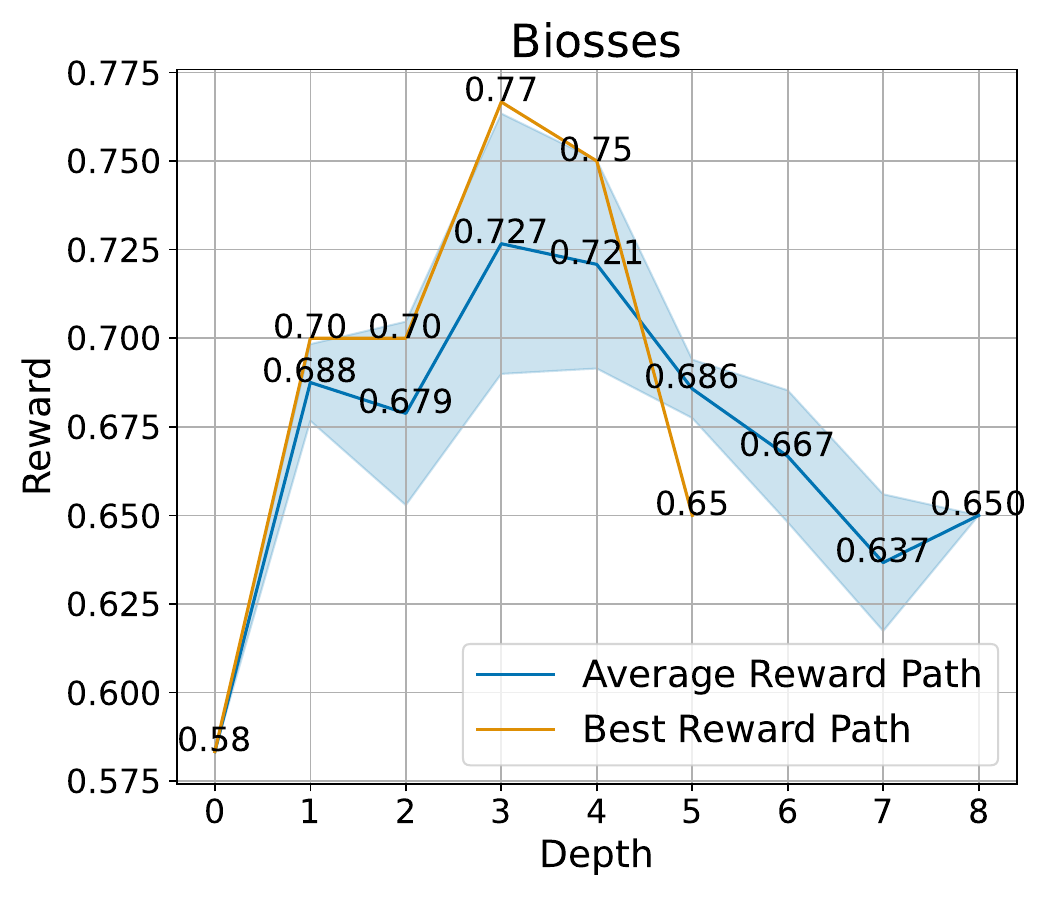}
    \end{subfigure}%
    \hspace{3em} 
    \begin{subfigure}{0.4\textwidth}
        \includegraphics[width=\textwidth]{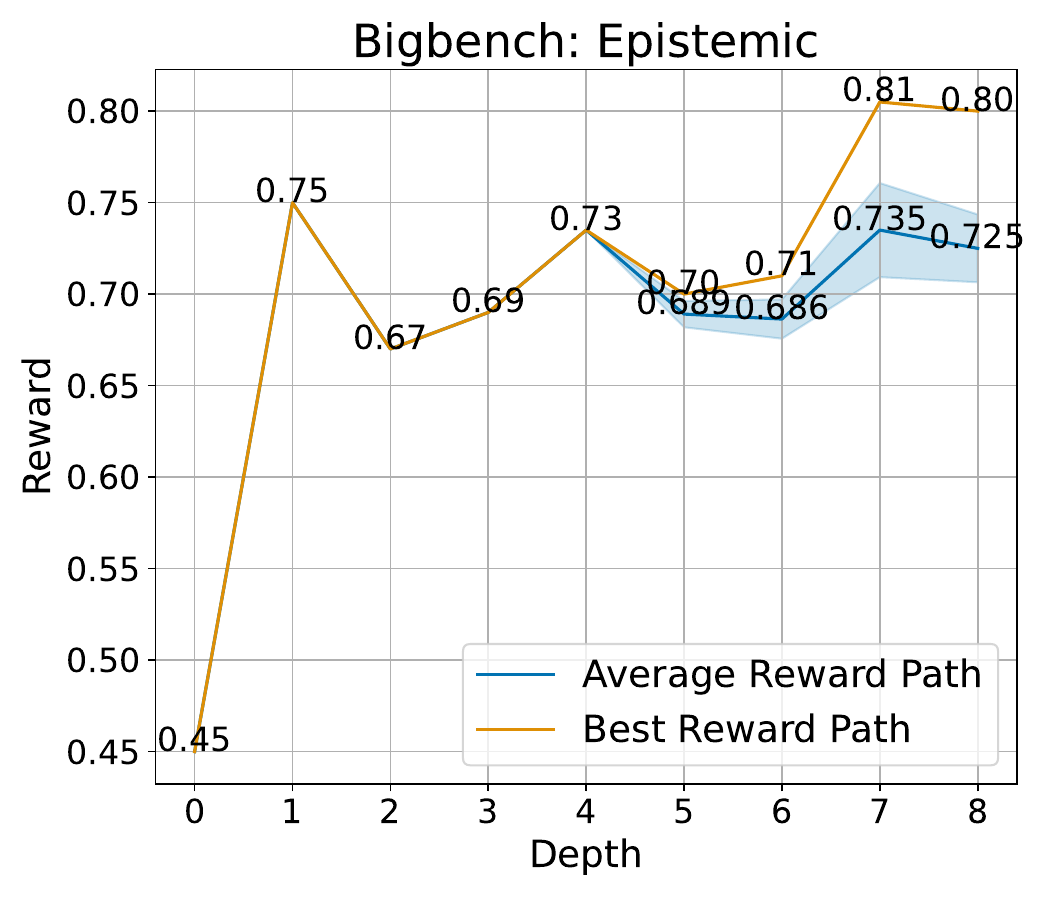}
    \end{subfigure}
    \vspace{0.6em} 

    \begin{subfigure}{0.4\textwidth}
        \includegraphics[width=\textwidth]{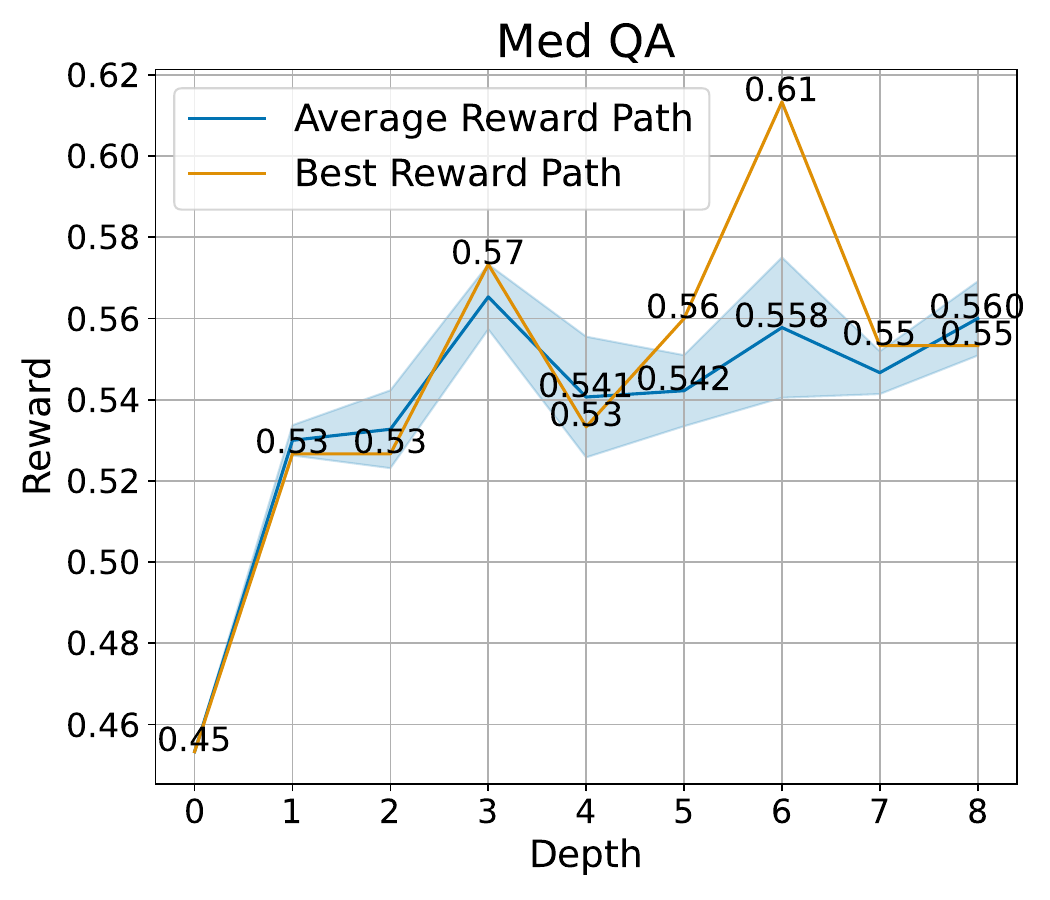}
    \end{subfigure}%
    \hspace{3em} 
    \begin{subfigure}{0.4\textwidth}
        \includegraphics[width=\textwidth]{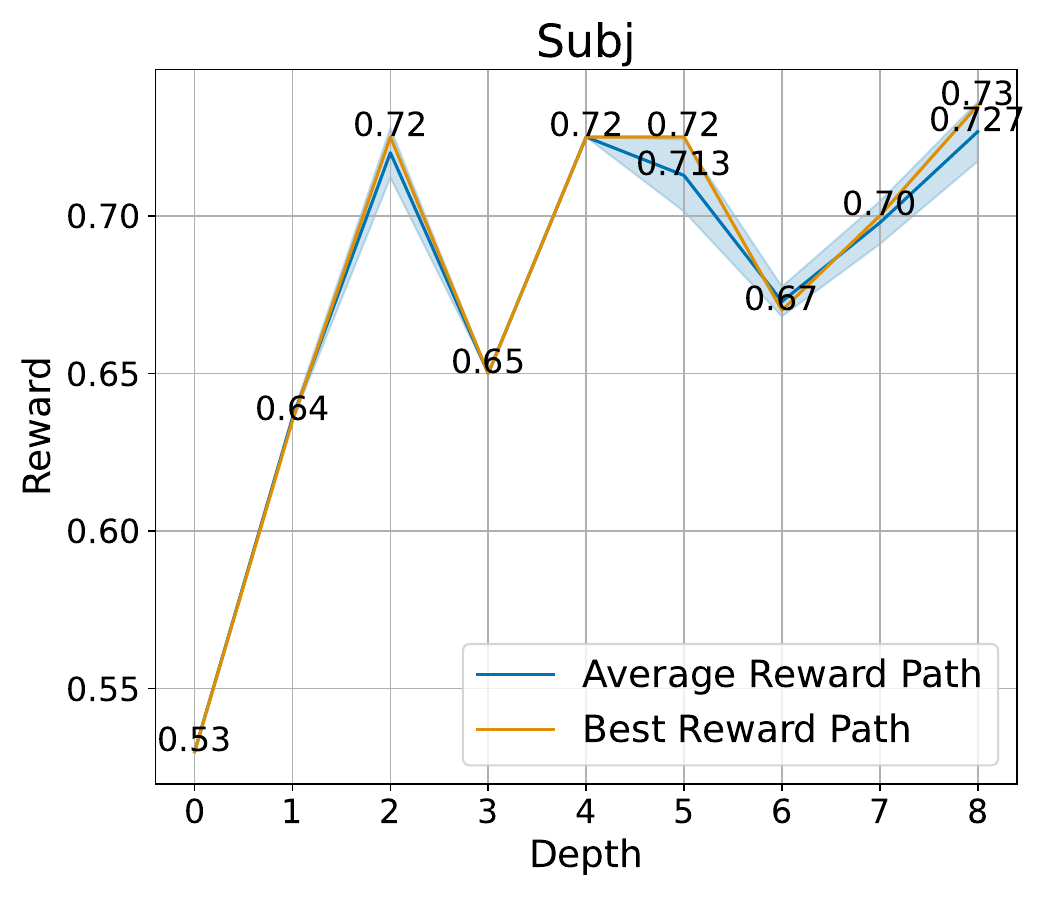}
    \end{subfigure}
    \vspace{0.6em} 

    \caption{Convergence plots with the ``Standard'' setting. $\textit{expand\_width}=3$, $\textit{num\_samples}=1$, and $\textit{depth\_limit}=8$. The Average Reward Path is the average reward of paths, and the blue area is the variance. The Best Reward Path is the path with highest average reward, where the best node is selected as the node with highest reward on the Best Reward Path.}
    \label{fig:mygraphs}
\end{figure}

\clearpage
\section{Optimized Prompts from \ours}
\label{sec:prompt_annotation}

In this section, we present the optimized prompt for all tasks, illustrating how \ours optimized prompts are different from ordinary human-written prompts and APE-optimized prompts.

\newcommand{\reduline}[1]{{\color{red}\underline{{\color{black}#1}}}}

\newcommand\dunderline[2][.2pt]{\raisebox{-#1}{\underline{\raisebox{#1}{\smash{\underline{#2}}}}}}

\begin{table}[h]
\caption{
Prompt comparison for the Geometric Shapes task, 
including normal human prompt, APE-optimized prompt, and expert-level prompt optimized by PromptAgent. Both baselines mostly describe the task, while our expert prompt is composed of more complex structures and domain-specific insights, achieving superior performance. {Bold text} denotes \textbf{domain knowledge} usually handcrafted by domain specialists, but here automatically discovered by \ours. We highlight different aspects of expert prompt with colors, including \ctext[RGB]{233,252,232}{Task Description}, \ctext[RGB]{255,230,230}{Term Clarification}, \ctext[RGB]{230,246,255}{Solution Guidance}, \ctext[RGB]{255,230,200}{Exception Handling}, \ctext[RGB]{255,225,255}{Priority \& Emphasis}, \ctext[RGB]{230, 230, 255}{Formatting}. (Best view with colors)
}
\label{tab: glues} 
\definecolor{Gray}{gray}{0.90}
\newcolumntype{a}{>{\columncolor{Gray}}c}
\centering
\resizebox{0.9\linewidth}{!}{%
\begin{tabular}{@{}lp{10cm}c@{}}
\toprule
Approach    & Optimized Prompt & Acc. \\ \midrule
Human       &     \ctext[RGB]{233,252,232}{Name geometric shapes from their SVG paths.}           &    0.227 \\\addlinespace
APE         &     \ctext[RGB]{233,252,232}{"Determine the shape each SVG path element is drawing, then pair it with the corresponding letter from the available choices. }\ctext[RGB]{255,230,230}{In this case, C symbolizes hexagon, G is for pentagon, I signifies sector, and B stands for heptagon."}      &     0.490 \\\addlinespace
PromptAgent &       
\ctext[RGB]{233,252,232}{In this task, you are tasked with interpreting SVG paths to determine the geometric figure they represent. }
{\ctext[RGB]{255,230,230}{The paths are delineated by commands: \textbf{'M' (move to), 'L' (line to), and 'A' (arc). An 'M' command initiates a path, potentially fragmenting a path into sub-paths, but it's crucial to not immediately view each 'M' as the starting point of a disconnected figure; often, they may continue the same geometric shape, manifesting as different sections within it. 'L' commands constitute line segments thus forming the boundaries of the figure. 'A' commands generate arcs, and depending on their sequence, can shape circles, sectors, elliptical figures, or other geometrical shapes through a continuous line of action. }}\ctext[RGB]{255,230,200}{\textbf{Note that an 'A' command followed by an 'L' could lead to specific shapes like sectors. }}}\ctext[RGB]{230,246,255}{Examine the sequence and interplay of 'M', 'L', and 'A' commands, as they together mold the final geometric figure and significantly govern its continuity. Potential shapes to be identified can range from simple lines to complex polygons. }{\ctext[RGB]{255,230,200}{'None of the above' is only a valid response if otherwise stated in the task. }}\ctext[RGB]{230,246,255}{As you formulate your answer, substantiate it with a clear explanation that encompasses the functionality of each command, their collective effect, sequence, and their correlational aspects. }\ctext[RGB]{255,230,200}{\textbf{In scenarios with multiple 'M' commands, refrain from arbitrarily breaking up the shape into disconnected figures; instead, visualize them contributing to different sections of the same shape. } } \ctext[RGB]{225,246,255}{Accurately count 'L' commands as they define the figure's sides, even when an 'M' command is present. For figuring out the entire geometric shape, meticulously examine all its components and commands, keeping an unbroken perception of the shape's progression, especially with multiple 'M' commands. }\ctext[RGB]{255,225,255}{Before finalizing your answer, recount the sides and arcs accurately - such a double-check ensures flawless identification of the geometric figure.}    & 0.670

\\ \bottomrule
\end{tabular}
}
\end{table}

\begin{table}[h]
\caption{
Prompt comparison for the Penguins In A Table task, 
including normal human prompt, APE-optimized prompt, and expert-level prompt optimized by PromptAgent. Both baselines mostly describe the task, while our expert prompt is composed of more complex structures and domain-specific insights, achieving superior performance. {Bold text} denotes \textbf{domain knowledge} usually handcrafted by domain specialists, but here automatically discovered by \ours. We highlight different aspects of expert prompt with colors, including \ctext[RGB]{233,252,232}{Task Description}, \ctext[RGB]{255,230,230}{Term Clarification}, \ctext[RGB]{230,246,255}{Solution Guidance}, \ctext[RGB]{255,230,200}{Exception Handling}, \ctext[RGB]{255,225,255}{Priority \& Emphasis}, \ctext[RGB]{230, 230, 255}{Formatting}. (Best view with colors)
}
\label{tab: glues} 
\definecolor{Gray}{gray}{0.90}
\newcolumntype{a}{>{\columncolor{Gray}}c}
\centering
\resizebox{0.9\linewidth}{!}{%
\begin{tabular}{@{}lp{10cm}c@{}}
\toprule
Approach    & Optimized Prompt & Acc. \\ \midrule
Human       &     \ctext[RGB]{233,252,232}{Answer questions about a table of penguins and their attributes.}          &    0.595\\\addlinespace
APE         &   \ctext[RGB]{233,252,232}{Carefully scrutinize the provided table or tables. Understand the query in relation to the information given. Pinpoint the pertinent data and carry out the vital computations or comparisons to determine the right answer from the given choices.}      &    0.747 \\\addlinespace
PromptAgent &       
\ctext[RGB]{230,246,255}{As you delve into a dataset of penguins, assess \textbf{essential attributes like names, ages, and gender}. Decode the significance of each attribute in the context of every penguin while \textbf{keeping in mind that the dataset may be modified, including addition or removal of penguins}. When such modifications are made, immediately revise your understanding, redo your computations, and ensure that your subsequent calculations consider these changes. }\ctext[RGB]{255,225,255}{The crux of your task is to identify relationships and patterns within the attributes, \textbf{giving special attention to the names and ages of the penguins}.}

\ctext[RGB]{230,246,255}{For complex tasks, break them down into manageable chunks ensuring no essential detail is missed. When a change is made to the dataset, recompute your values taking into consideration these changes, paying extra attention to cumulative computations. }\ctext[RGB]{255,230,230}{\textbf{Ensure that your understanding of 'more than', 'less than', and 'equal to' is precise and that you correctly interpret these in context of the question. }}  

\ctext[RGB]{230,246,255}{Put into place a verification mechanism to authenticate the accuracy of your solutions, stating out your understanding of the query and the assumptions you have made to resolve it. }\ctext[RGB]{255,225,255}{\textbf{Bear in mind that tasks may require you to combine the dataset with additional external information, this may include understanding age disparities outside explicit lifespan parameters, identifying common names linked to gender, or recognizing names associated with famous individuals. }Document your matters of interest meticulously and maintain rigorous accuracy levels in your calculations to prevent errors.}

\ctext[RGB]{230,246,255}{Stay nimble-footed in reshaping your analytical approach based on each new query.  This might include uncovering numerical patterns, comprehending inherent data natures, or liaising with external sources for a more thorough understanding. }\ctext[RGB]{255,225,255}{\textbf{Most importantly, prior to making a comparison within attributes such as age or height, conduct a thorough investigation of all values under that attribute.}}\ctext[RGB]{230,246,255}{Understand the premise of each question before springing to deductions, and remember, any change in the dataset denotes a new starting point for the following computational steps to maintain accuracy.}     &0.873

\\ \bottomrule
\end{tabular}
}
\end{table}

\begin{table}[h]
\caption{
Prompt comparison for the Epistemic Reasoning task, 
including normal human prompt, APE-optimized prompt, and expert-level prompt optimized by PromptAgent. Both baselines mostly describe the task, while our expert prompt is composed of more complex structures and domain-specific insights, achieving superior performance. {Bold text} denotes \textbf{domain knowledge} usually handcrafted by domain specialists, but here automatically discovered by \ours. We highlight different aspects of expert prompt with colors, including \ctext[RGB]{233,252,232}{Task Description}, \ctext[RGB]{255,230,230}{Term Clarification}, \ctext[RGB]{230,246,255}{Solution Guidance}, \ctext[RGB]{255,230,200}{Exception Handling}, \ctext[RGB]{255,225,255}{Priority \& Emphasis}, \ctext[RGB]{230, 230, 255}{Formatting}. (Best view with colors)
}
\label{tab: glues} 
\definecolor{Gray}{gray}{0.90}
\newcolumntype{a}{>{\columncolor{Gray}}c}
\centering
\resizebox{0.9\linewidth}{!}{%
\begin{tabular}{@{}lp{10cm}c@{}}
\toprule
Approach    & Optimized Prompt & Acc. \\ \midrule
Human       &      \ctext[RGB]{233,252,232}{Determine whether one sentence entails the next. }          &    0.452 \\\addlinespace
APE         &      \ctext[RGB]{233,252,232}{Determine whether the hypothesis is directly implied by the premise or not. }\ctext[RGB]{255,230,230}{If the premise's statement is a direct claim or conviction of the individual mentioned in the hypothesis, choose 'entailment'. However, if the premise is formed on the belief or supposition of someone other than the subject in the hypothesis, opt for 'non-entailment'.}       &     0.708 \\\addlinespace
PromptAgent &       
\ctext[RGB]{233,252,232}{Your task is to critically analyse the primary sentence, known as the 'premise', with the objective of determining whether it unequivocally supports the truth value of the subsequent sentence or 'hypothesis'.}\ctext[RGB]{230, 230, 255}{ The relationship between the premise and hypothesis can be classified as 'Entailment' or 'Non-Entailment'.  }\ctext[RGB]{255,230,230}{Label it as 'Entailment' if the premise provides \textbf{robust evidence substantiating the truth of the hypothesis without requiring additional context}. If, however, the corroboration of the hypothesis by the premise is not entirely explicit, select 'Non-Entailment'.}

\ctext[RGB]{230,246,255}{Deciphering the semantics within the sentences is crucial for your final decision.  }\ctext[RGB]{255,230,230}{\textbf{Terms such as 'assumes', 'believes', 'thinks', 'feels', 'suspects', and their likes should be respected for their capacity to introduce uncertainty and subjectivity, and not perceived as conclusive proof of the hypothesis, regardless of whether they form part of nested beliefs or not.} }\ctext[RGB]{255,230,200}{\textbf{Also, a detailed premise does not necessarily negate a more generalized hypothesis.  For example, a premise that mentions a 'full face mask' correlates to a hypothesis that states a 'mask'.}}

\ctext[RGB]{255,225,255}{During your evaluation, maintain a keen focus on factual and logical reasoning, always bearing in mind that personal beliefs or experiences should be incorporated into your review only if they are inherently connected to the factual content of the statements.  }\ctext[RGB]{255,230,200}{\textbf{However, these should be understood as subjective truths in the context of the individual's perspective and should not be taken as objectively verifiable truths.}}

\ctext[RGB]{230,246,255}{Upon deciding between 'Entailment' or 'Non-Entailment', articulate your explanations in a concise manner, warranting that you desist from making precipitous conclusions or unsupported assumptions. Your judgement should be firmly anchored in the logical and factual ties existing within the premise and hypothesis, renouncing any incidental inferences or personal interpretations.

\textbf{Exercise restraint in passing verdicts on the truth value or validity of personal beliefs, unless they have a direct bearing on the factual correlation between the premise and the hypothesis}. During your estimation, mindfully weigh the extent of uncertainty introduced by expressions of belief or suspicion against the imperative for factual precision when establishing the entailment.}     & 0.806

\\ \bottomrule
\end{tabular}
}
\end{table}

\begin{table}[h]
\caption{
Prompt comparison for the Object Counting task, 
including normal human prompt, APE-optimized prompt, and expert-level prompt optimized by PromptAgent. Both baselines mostly describe the task, while our expert prompt is composed of more complex structures and domain-specific insights, achieving superior performance. {Bold text} denotes \textbf{domain knowledge} usually handcrafted by domain specialists, but here automatically discovered by \ours. We highlight different aspects of expert prompt with colors, including \ctext[RGB]{233,252,232}{Task Description}, \ctext[RGB]{255,230,230}{Term Clarification}, \ctext[RGB]{230,246,255}{Solution Guidance}, \ctext[RGB]{255,230,200}{Exception Handling}, \ctext[RGB]{255,225,255}{Priority \& Emphasis}, \ctext[RGB]{230, 230, 255}{Formatting}. (Best view with colors)
}
\label{tab: glues} 
\definecolor{Gray}{gray}{0.90}
\newcolumntype{a}{>{\columncolor{Gray}}c}
\centering
\resizebox{0.9\linewidth}{!}{%
\begin{tabular}{@{}lp{10cm}c@{}}
\toprule
Approach    & Optimized Prompt & Acc. \\ \midrule
Human       &     \ctext[RGB]{233,252,232}{Count the overall number of all items.}           &    0.612\\\addlinespace
APE         &   \ctext[RGB]{233,252,232}{Calculate the overall total of all items even those spoken in groups.}      &     0.716 \\\addlinespace
PromptAgent &       
\ctext[RGB]{233,252,232}{Carefully analyze the given information. Catalog each item mentioned and denote any explicitly defined quantities. }\ctext[RGB]{255,230,200}{\textbf{If an item - quantity is not stated, assume it as a single unit. However, for an item with a specified quantity, make sure to count each unit separately and include it in your total count.}}\ctext[RGB]{230,246,255}{ If collective terms or categories are identified, break them down into their individual components and reasonably associate each with its stated count. Proceed to calculate a comprehensive total for such categories ensuring the sum includes all individual units, not the number of subsets or types. }\ctext[RGB]{255,225,255}{\textbf{Remember that each item has its unique count, but items related or falling under a common category should be tabulated as such, with their individual quantities precisely contributing to the final count. }}\ctext[RGB]{230,246,255}{Avoid making assumptions about the nature or categorization of items and adhere to commonly accepted definitions and classifications. Review your work to ensure accuracy and to avoid mistakes in counting. \textbf{Modify your strategy if required by considering items within varying categories, types, or subtypes}. Eventually, summarize the count indicating the specific quantity for each identified item or category and a total count of units, not categories, or provide a comprehensive overview as explicitly requested.}     & 0.86

\\ \bottomrule
\end{tabular}
}
\end{table}

\begin{table}[h]
\caption{
Prompt comparison for the Temporal Sequences task, 
including normal human prompt, APE-optimized prompt, and expert-level prompt optimized by PromptAgent. Both baselines mostly describe the task, while our expert prompt is composed of more complex structures and domain-specific insights, achieving superior performance. {Bold text} denotes \textbf{domain knowledge} usually handcrafted by domain specialists, but here automatically discovered by \ours. We highlight different aspects of expert prompt with colors, including \ctext[RGB]{233,252,232}{Task Description}, \ctext[RGB]{255,230,230}{Term Clarification}, \ctext[RGB]{230,246,255}{Solution Guidance}, \ctext[RGB]{255,230,200}{Exception Handling}, \ctext[RGB]{255,225,255}{Priority \& Emphasis}, \ctext[RGB]{230, 230, 255}{Formatting}. (Best view with colors)
}
\label{tab: glues} 
\definecolor{Gray}{gray}{0.90}
\newcolumntype{a}{>{\columncolor{Gray}}c}
\centering
\resizebox{0.9\linewidth}{!}{%
\begin{tabular}{@{}lp{10cm}c@{}}
\toprule
Approach    & Optimized Prompt & Acc. \\ \midrule
Human       &     \ctext[RGB]{233,252,232}{Answer questions about which times certain events could have occurred.}         &    0.72\\\addlinespace
APE         &  \ctext[RGB]{233,252,232}{Identify the period when the individual was unnoticed and had the possibility to visit the specified place before its closing time.}      &    0.856 \\\addlinespace
PromptAgent &       
\ctext[RGB]{233,252,232}{By examining the series of daily activities of an individual, pinpoint when they were free and when they were busy. }\ctext[RGB]{230,246,255}{Use these open slots to dictate when they could possibly engage in other activities. }\ctext[RGB]{255,230,200}{\textbf{Upon waking up, a person does not instantly become occupied. }}\ctext[RGB]{230,246,255}{\textbf{Take into account any potential restrictions or closed times and use these as an indicator that the event cannot take place during these hours. An overlap of activities is unallowable, so ensure there is no overlap while creating a timeline.} Cross-check the free time slots with the functioning hours of the potential event to accurately derive the most likely time interval for the event to take place.}     &0.934

\\ \bottomrule
\end{tabular}
}
\end{table}

\begin{table}[h]
\caption{ 
Prompt comparison for the Causal Judgment task, 
including normal human prompt, APE-optimized prompt, and expert-level prompt optimized by PromptAgent. Both baselines mostly describe the task, while our expert prompt is composed of more complex structures and domain-specific insights, achieving superior performance. {Bold text} denotes \textbf{domain knowledge} usually handcrafted by domain specialists, but here automatically discovered by \ours. We highlight different aspects of expert prompt with colors, including \ctext[RGB]{233,252,232}{Task Description}, \ctext[RGB]{255,230,230}{Term Clarification}, \ctext[RGB]{230,246,255}{Solution Guidance}, \ctext[RGB]{255,230,200}{Exception Handling}, \ctext[RGB]{255,225,255}{Priority \& Emphasis}, \ctext[RGB]{230, 230, 255}{Formatting}. (Best view with colors)
}
\label{tab: glues} 
\definecolor{Gray}{gray}{0.90}
\newcolumntype{a}{>{\columncolor{Gray}}c}
\centering
\resizebox{0.9\linewidth}{!}{%
\begin{tabular}{@{}lp{10cm}c@{}}
\toprule
Approach    & Optimized Prompt & Acc. \\ \midrule
Human       &      \ctext[RGB]{233,252,232}{Answer questions about causal attribution.}            &    0.47  \\\addlinespace
APE         &       \ctext[RGB]{233,252,232}{"For each situation, decide if the result was caused deliberately or not. }\ctext[RGB]{255,230,230}{If the individual or party behind the event was aware of the potential result and chose to go ahead, select 'A'. If they didn't intend the result to happen, even if they knew it could possibly occur, select 'B'."}        &     0.57 \\\addlinespace
PromptAgent &       
\ctext[RGB]{233,252,232}{Respond to inquiries about causal attribution, focusing on the entity or entities specifically highlighted in the question. }\ctext[RGB]{230,246,255}{\textbf{Carefully investigate multi-factorial causes that may operate simultaneously and independently}, and discern the underlying intentions behind an individual's actions. Differentiate between immediate and incidental origins and identify the contribution of each factor in creating the outcome. \textbf{Examine the interplay of causes within the immediate situation and larger systemic frameworks}. Maintain uncompromising adherence to the details provided within the context and restrain from making assumptions unsupported by the evidence presented. \textbf{Always consider the complexity of multiple causes contributing to a single effect and resist attributing the effect to a singular cause. Recognize the possibility of synergy amongst causes and its resultant effects}.}     & 0.67

\\ \bottomrule
\end{tabular}
}
\end{table}

\begin{table}[h]
\caption{ 
Prompt comparison for the Biosses task, 
including normal human prompt, APE-optimized prompt, and expert-level prompt optimized by PromptAgent. Both baselines mostly describe the task, while our expert prompt is composed of more complex structures and domain-specific insights, achieving superior performance. {Bold text} denotes \textbf{domain knowledge} usually handcrafted by domain specialists, but here automatically discovered by \ours. We highlight different aspects of expert prompt with colors, including \ctext[RGB]{233,252,232}{Task Description}, \ctext[RGB]{255,230,230}{Term Clarification}, \ctext[RGB]{230,246,255}{Solution Guidance}, \ctext[RGB]{255,230,200}{Exception Handling}, \ctext[RGB]{255,225,255}{Priority \& Emphasis}, \ctext[RGB]{230, 230, 255}{Formatting}. (Best view with colors)
}
\label{tab: glues} 
\definecolor{Gray}{gray}{0.90}
\newcolumntype{a}{>{\columncolor{Gray}}c}
\centering
\resizebox{0.9\linewidth}{!}{%
\begin{tabular}{@{}lp{10cm}c@{}}
\toprule
Approach    & Optimized Prompt & Acc. \\ \midrule
Human       &      \ctext[RGB]{233,252,232}{This is a biomedical sentence similarity task. Please carefully read the following sentences and rate the similarity of two input sentences. }\ctext[RGB]{230, 230, 255}{Choose between 'not similar', 'somewhat similar' and 'similar'}            &    0.55  \\\addlinespace
APE         &       \ctext[RGB]{233,252,232}{"Examine the two given sentences and assess their content similarity. }\ctext[RGB]{255,230,230}{Choice A (not similar) should be selected if the sentences discuss entirely different topics or concepts. Choose option B (somewhat similar) if they have some common points but also contain differences. Select option C (similar) if the sentences primarily convey the same message or could be used in place of one another."}        &     0.7 \\\addlinespace
PromptAgent &       
\ctext[RGB]{233,252,232}{For this task, you are asked to perform a biomedical sentence similarity evaluation. }\ctext[RGB]{230,246,255}{Examine the two input sentences and evaluate their similarity, not only taking into account common terms or concepts \textbf{but also the complex scientific language, specific processes, and unique subject matter they delve into}. \textbf{Consider not only the subject matter but also the intended purpose like whether they both describe a process, report a finding, or detail a method or technique}. }\ctext[RGB]{255,230,230}{Rate the similarity as 'not similar' if their subject matter or emphasis is distinct, 'somewhat similar' if they discuss related topics or share some details but are not entirely identical, and 'similar' if the sentences precisely mirror each other in topic and conclusions. }\ctext[RGB]{255,225,255}{Remember, this task requires more than a cursory scan of keywords - focus on the nuanced meanings, pay attention to the degree at which the discussed concepts or processes are general or specific, and strive for a comprehensive understanding of the contents.}     & 0.75

\\ \bottomrule
\end{tabular}
}
\end{table}

\begin{table}[h]
\caption{
Prompt comparison for the Med\_QA task, 
including normal human prompt, APE-optimized prompt, and expert-level prompt optimized by PromptAgent. Both baselines mostly describe the task, while our expert prompt is composed of more complex structures and domain-specific insights, achieving superior performance. {Bold text} denotes \textbf{domain knowledge} usually handcrafted by domain specialists, but here automatically discovered by \ours. We highlight different aspects of expert prompt with colors, including \ctext[RGB]{233,252,232}{Task Description}, \ctext[RGB]{255,230,230}{Term Clarification}, \ctext[RGB]{230,246,255}{Solution Guidance}, \ctext[RGB]{255,230,200}{Exception Handling}, \ctext[RGB]{255,225,255}{Priority \& Emphasis}, \ctext[RGB]{230, 230, 255}{Formatting}. (Best view with colors)
}
\label{tab: glues} 
\definecolor{Gray}{gray}{0.90}
\newcolumntype{a}{>{\columncolor{Gray}}c}
\centering
\resizebox{0.9\linewidth}{!}{%
\begin{tabular}{@{}lp{10cm}c@{}}
\toprule
Approach    & Optimized Prompt & Acc. \\ \midrule
Human       &     \ctext[RGB]{233,252,232}{Please use your domain knowledge in medical area to solve the questions.}           &    0.508 \\\addlinespace
APE         &    \ctext[RGB]{233,252,232}{"For every presented clinical situation, scrutinize the symptoms and specifics given. }\ctext[RGB]{230, 230, 255}{From the options A-E, choose the one that best pinpoints the cause or diagnosis of the stated condition."}      &     0.47 \\\addlinespace
PromptAgent &       
\ctext[RGB]{233,252,232}{Leveraging particularly your comprehensive medical expertise, handle each presented scenario as you would a complicated puzzle requiring careful, unbiased assessment. }\ctext[RGB]{255,225,255}{\textbf{Each nugget of information - from patient age, gender, lifestyle, symptoms, lab results, and past medical history}, to recent activities that may be relevant to their condition, plays an equally important role in shaping your judgement.}

\ctext[RGB]{255,230,200}{\textbf{Becoming cognizant of the fact that medical conditions can manifest uniquely in different individuals is crucial; avoid precipitating conclusions merely on the basis of stereotypical symptoms}. }\ctext[RGB]{230,246,255}{Instead, employ a deep understanding of the variety of medical conditions to critically evaluate each symptom's relevance, ensuring that undue bias is not allocated to particular symptoms over others.} 

\ctext[RGB]{255,225,255}{Particularly, pay attention to common symptoms over rare ones unless otherwise indicated. }\ctext[RGB]{230,246,255}{Break down assumptions and consider the most likely cause in a given context. \textbf{Do not overlook the importance of demographic details and their correlation with symptoms, especially when a symptom hints at a particular physiological state, like menopause}.

Through meticulous examination, ensure you grasp the nuances in each query's context, \textbf{with keen focus on the developmental stages in children and the specific challenges they entail}. Capture the timelines of symptoms, understanding that often, a diagnosis relies significantly on the onset and duration of these symptoms.

Once conclusions begin taking shape, undertake an exhaustive cross-verification exercise with the available multiple choice answers. Evaluate these options for relevance and decide their probability on the specifics of the given case. Abstain from dismissing potential answers at first glance, but rather advocate for an intensive assessment of all.

Approach scenarios similar to solving a complex jigsaw puzzle. \textbf{Each distinct symptom, lab result, past medical history, and timing forms an integral component that lends weight to a deeper comprehension of the patient's present condition}. The endgame extends beyond merely achieving precision and a comprehensive enquiry but ensures that your conclusions do not yield overgeneralization or oversimplification towards the diagnosis and treatment therein.

\textbf{Examine closely every symptom in relation to the disease and differentiate those that are side effects of treatment. Be cautious when multiple symptoms present simultaneously, to avoid confusion}. The imprint of your insight should reflect a holistic understanding of the case, zooming into the most probable diagnosis or treatment strategy that suits the breadth of data at disposal.}     & 0.57

\\ \bottomrule
\end{tabular}
}
\end{table}

\begin{table}[h]
\caption{ 
Prompt comparison for the Subjective task, 
including normal human prompt, APE-optimized prompt, and expert-level prompt optimized by PromptAgent. Both baselines mostly describe the task, while our expert prompt is composed of more complex structures and domain-specific insights, achieving superior performance. {Bold text} denotes \textbf{domain knowledge} usually handcrafted by domain specialists, but here automatically discovered by \ours. We highlight different aspects of expert prompt with colors, including \ctext[RGB]{233,252,232}{Task Description}, \ctext[RGB]{255,230,230}{Term Clarification}, \ctext[RGB]{230,246,255}{Solution Guidance}, \ctext[RGB]{255,230,200}{Exception Handling}, \ctext[RGB]{255,225,255}{Priority \& Emphasis}, \ctext[RGB]{230, 230, 255}{Formatting}. (Best view with colors)
}
\label{tab: glues} 
\definecolor{Gray}{gray}{0.90}
\newcolumntype{a}{>{\columncolor{Gray}}c}
\centering
\resizebox{0.9\linewidth}{!}{%
\begin{tabular}{@{}lp{10cm}c@{}}
\toprule
Approach    & Optimized Prompt & Acc. \\ \midrule
Human       &    \ctext[RGB]{233,252,232}{Given the text, choose between 'subjective' and 'objective'.}         &    0.517\\\addlinespace
APE         &     \ctext[RGB]{233,252,232}{Determine whether the provided text is stating }\ctext[RGB]{255,230,230}{facts and details (Objective) or expressing personal views, emotions, or choices (Subjective).}      &     0.696 \\\addlinespace
PromptAgent & \ctext[RGB]{233,252,232}{Examine the given text and decide whether it is 'subjective' or 'objective'. }\ctext[RGB]{255,230,230}{Define the narrative as 'subjective' \textbf{if it seems to be significantly swayed by the author's personal emotions, viewpoints, or beliefs}. Conversely, 'objective' narratives should \textbf{impartially depict facts or scenarios, devoid of personal prejudices, preconceived beliefs, and the author's own convictions}.}\ctext[RGB]{255,230,200}{ \textbf{It is essential to understand that emotionally-dense language, vivid descriptions or depiction of characters' emotional states do not always hint at subjectivity}. They may just serve to represent situations authentically without conveying the author's personal standpoint. \textbf{Unconventional punctuation, dialogues or queries do not inherently contribute to authorial subjectivity}. }\ctext[RGB]{230,246,255}{Draw a clear distinction between the author's and characters' subjectivity; misinterpreting a character's subjectivity as the author's personal bias is a common pitfall.} \ctext[RGB]{255,225,255}{The priority is to extract the author's tendency within the narrative, rather than focusing on the characters. Utilize these directives to critically analyze the text.}  & 0.806

\\ \bottomrule
\end{tabular}
}
\end{table}

\begin{table}[h]
\caption{
Prompt comparison for the TREC task, 
including normal human prompt, APE-optimized prompt, and expert-level prompt optimized by PromptAgent. Both baselines mostly describe the task, while our expert prompt is composed of more complex structures and domain-specific insights, achieving superior performance. {Bold text} denotes \textbf{domain knowledge} usually handcrafted by domain specialists, but here automatically discovered by \ours. We highlight different aspects of expert prompt with colors, including \ctext[RGB]{233,252,232}{Task Description}, \ctext[RGB]{255,230,230}{Term Clarification}, \ctext[RGB]{230,246,255}{Solution Guidance}, \ctext[RGB]{255,230,200}{Exception Handling}, \ctext[RGB]{255,225,255}{Priority \& Emphasis}, \ctext[RGB]{230, 230, 255}{Formatting}. (Best view with colors)
}
\label{tab: glues} 
\definecolor{Gray}{gray}{0.90}
\newcolumntype{a}{>{\columncolor{Gray}}c}
\centering
\resizebox{0.9\linewidth}{!}{%
\begin{tabular}{@{}lp{10cm}c@{}}
\toprule
Approach    & Optimized Prompt & Acc. \\ \midrule
Human       &    \ctext[RGB]{233,252,232}{Tag the text according to the primary topic of the question. Choose from (A) Abbreviation, (B) Entity, (C) Description and abstract concept, (D) Human being, (E) Location, (F) Numeric value}         &    0.742\\\addlinespace
APE         &     \ctext[RGB]{233,252,232}{"Tag the text according to the primary topic of the question. }\ctext[RGB]{255,230,230}{Select 'Human being' (D) if the question \textbf{revolves around a person}. Opt for 'Description and abstract concept' (C) if the question \textbf{requires an explanation or description of a concept}. Choose 'Location' (E) if the question is about a specific place. If the question refers to a particular object or thing, then select 'Entity' (B). If the question \textbf{involves data or a length of time}, opt for 'Numeric value' (F). Disregard 'Abbreviation' (A) since it's not related to any of the questions."}    &     0.834 \\\addlinespace
PromptAgent &       
\ctext[RGB]{233,252,232}{For the question given above, determine the type of response it is aiming to elicit, then assign the most fitting label from the following: }\ctext[RGB]{255,230,230}{(A) Abbreviation, (B) Tangible and Intangible Entity \textbf{(including distinct terms, theories, inventions, phenomena)}, (C) Description and Abstract Concept \textbf{(concerning explanations, clarifications, theoretical ideas)}, (D) Individual and Collective Humans \textbf{(encompassing distinct persons, the creators of certain works, groups, organizations)}, (E) Location, or (F) Numeric Value \textbf{(containing numeric figures, dates, timings, quantities)}.}\ctext[RGB]{255,225,255}{ The key is the answer-type the question is seeking, not other elements in the question. Your assigned label should prioritize the primary response over additional details. }\ctext[RGB]{230,246,255}{If a solo label does not closely address the entire answer intent of the question, then you may assign more than one. The label should reflect the assumed answer's nature, not the mere question's content or incidental features. Place the label you consider most fitting for the question's main intention.}  & 0.886

\\ \bottomrule
\end{tabular}
}
\end{table}

\begin{table}[h]
\caption{
Prompt comparison for the CB task, 
including normal human prompt, APE-optimized prompt, and expert-level prompt optimized by PromptAgent. Both baselines mostly describe the task, while our expert prompt is composed of more complex structures and domain-specific insights, achieving superior performance. {Bold text} denotes \textbf{domain knowledge} usually handcrafted by domain specialists, but here automatically discovered by \ours. We highlight different aspects of expert prompt with colors, including \ctext[RGB]{233,252,232}{Task Description}, \ctext[RGB]{255,230,230}{Term Clarification}, \ctext[RGB]{230,246,255}{Solution Guidance}, \ctext[RGB]{255,230,200}{Exception Handling}, \ctext[RGB]{255,225,255}{Priority \& Emphasis}, \ctext[RGB]{230, 230, 255}{Formatting}. (Best view with colors)
}
\label{tab: glues} 
\definecolor{Gray}{gray}{0.90}
\newcolumntype{a}{>{\columncolor{Gray}}c}
\centering
\resizebox{0.9\linewidth}{!}{%
\begin{tabular}{@{}lp{10cm}c@{}}
\toprule
Approach    & Optimized Prompt & Acc. \\ \midrule
Human       &      \ctext[RGB]{233,252,232}{Read carefully the following premise and hypothesis, and determine the relationship between them. Choose from 'contradiction', 'neutral' and 'entailment'}.           &    0.714\\\addlinespace
APE         &      "\ctext[RGB]{233,252,232}{Ascertain the link between the premise and the hypothesis.} \ctext[RGB]{255,230,230}{If the hypothesis happens to be a rational outcome or inference from the premise, label it as an 'entailment'. If the hypothesis presents a contrasting scenario or clashes with the premise, categorize it as a 'contradiction'. In case the hypothesis neither disputes nor is it derived from the premise, term it as 'neutral'.}"       &     0.8036 \\\addlinespace
PromptAgent &       
\ctext[RGB]{233,252,232}{Your task is to delve deeply into the provided premise and hypothesis. }\ctext[RGB]{230,246,255}{Highlight explicit, central information and important entities mentioned in the dialogue while considering multiple ways the same thought could be delivered through language. }\ctext[RGB]{255,230,230}{\textbf{Acknowledge that a hypothesis might reflect, rephrase, or reiterate ideas from the premise, possibly in a simplified manner.} }\ctext[RGB]{255,230,200}{However, remember that \textbf{mere verbatim repetition does not automatically signal 'entailment'.} }\ctext[RGB]{255,230,230}{The reiteration in the hypothesis should represent a pivotal idea in the premise for it to be categorized as entailment. \textbf{If the hypothesis asserts something diametrically opposed to what's stated in the premise, mark it as a 'contradiction'. Reserve 'neutral' for scenarios where the premise and the hypothesis appear disconnected or do not exhibit any clear relationship.} }\ctext[RGB]{230,246,255}{Be vigilant while dealing with ambiguities, and strive to decode them in the context of the hypothesis. Do not allow nuanced or hypothetical statements distract from identifying the primary idea in the hypothesis. }\ctext[RGB]{230, 230, 255}{Know that your classifications, 'entailment', 'contradiction', or 'neutral',}\ctext[RGB]{230,246,255}{ should mirror the essential relationship derived strictly from the premise and the hypothesis, without the influence of personal opinions or conclusions. }\ctext[RGB]{255,225,255}{Prioritize understanding the core intention and context of the conversation over mere repetition of words or phrases.}    & 0.911

\\ \bottomrule
\end{tabular}
}
\end{table}

\end{document}